\documentclass[10pt,journal]{IEEEtran}
\usepackage{cite}
\usepackage[pdftex]{graphicx}
\usepackage{amsmath,amsfonts}

\usepackage{url}
\usepackage{subcaption}
\usepackage[hidelinks]{hyperref}
\usepackage{xcolor}

\begin{document}
\bstctlcite{IEEEexample:BSTcontrol}

\title{Efficient High-Resolution Deep Learning: A Survey}

\author{Arian~Bakhtiarnia,
        Qi~Zhang,
        and~Alexandros~Iosifidis,
\thanks{Arian Bakhtiarnia, Qi Zhang and Alexandros Iosifidis are with DIGIT, the Department of Electrical and Computer Engineering, Aarhus University, Aarhus, Midtjylland, Denmark (e-mail: arianbakh@ece.au.dk; qz@ece.au.dk; ai@ece.au.dk).}%
\thanks{This work was funded by the European Union’s Horizon 2020 research and innovation programme under grant agreement No 957337, and by the Danish Council for Independent Research under Grant No. 9131-00119B. 
}}


\maketitle

\begin{abstract}
Cameras in modern devices such as smartphones, satellites and medical equipment are capable of capturing very high resolution images and videos. Such high-resolution data often need to be processed by deep learning models for cancer detection, automated road navigation, weather prediction, surveillance, optimizing agricultural processes and many other applications. Using high-resolution images and videos as direct inputs for deep learning models creates many challenges due to their high number of parameters, computation cost, inference latency and GPU memory consumption. Simple approaches such as resizing the images to a lower resolution are common in the literature, however, they typically significantly decrease accuracy. Several works in the literature propose better alternatives in order to deal with the challenges of high-resolution data and improve accuracy and speed while complying with hardware limitations and time restrictions. This survey describes such efficient high-resolution deep learning methods, summarizes real-world applications of high-resolution deep learning, and provides comprehensive information about available high-resolution datasets.
\end{abstract}

\section{Introduction}
\begin{figure*}
\begin{center}
\begin{tabular}{ c c }
\includegraphics[width=0.48\textwidth]{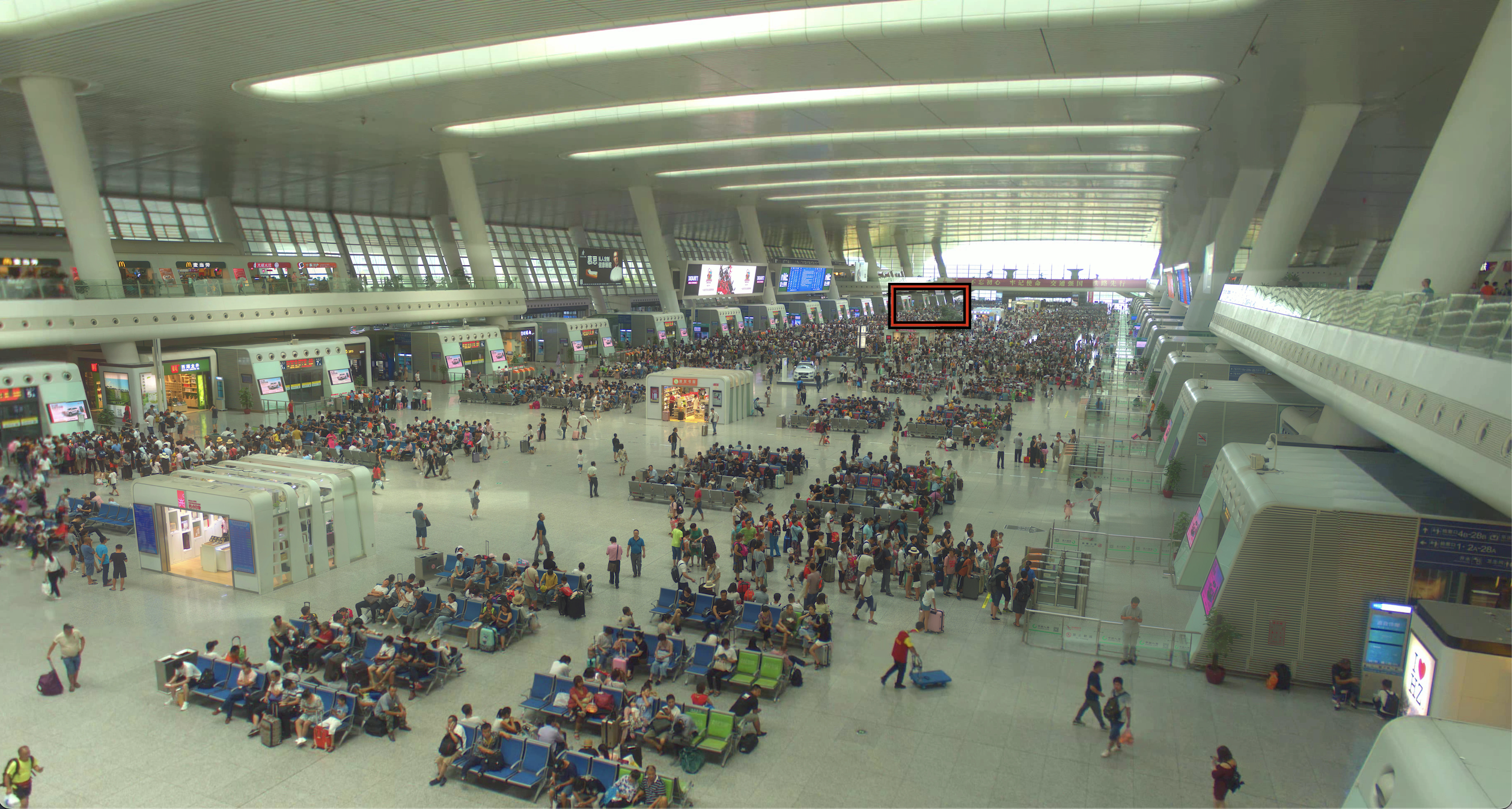} & \includegraphics[width=0.48\textwidth]{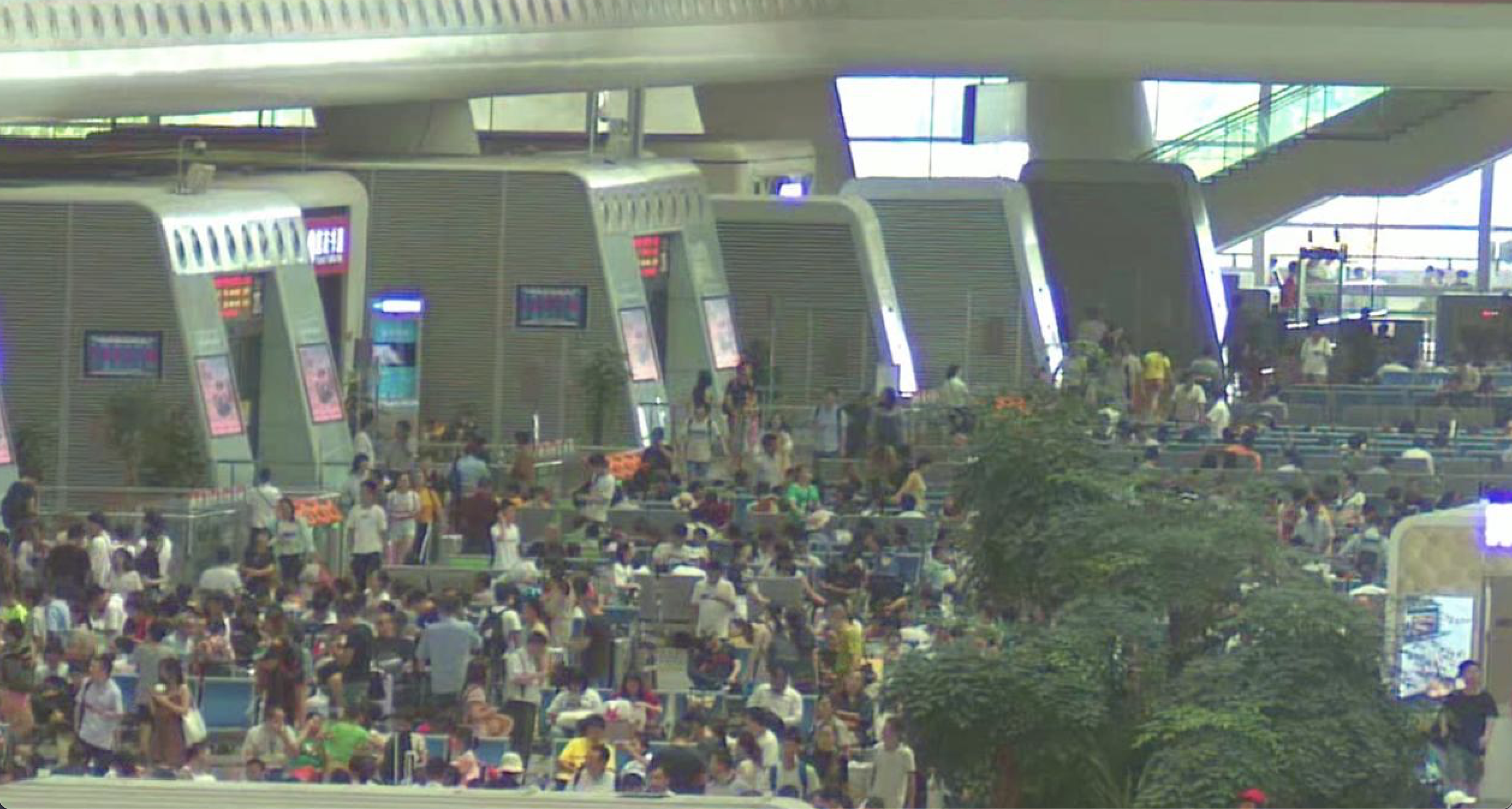}\\
(a) & (b) \\
\end{tabular}
\end{center}
\caption{Example of a gigapixel image, taken from the PANDA-Crowd dataset \cite{wang2020panda}, captured using an array of micro-cameras; (a) original image with a size of 26,558$ \times $14,828 pixels, and (b) zoomed in to the location specified by the red rectangle in the original image, with a size of 2,516$ \times $1,347 pixels, which is more than 100 times smaller than the original image, yet still approximately 5 times larger than the image size processed by state-of-the-art deep learning models for crowd counting such as SASNet \cite{song2021choose}, which is 1024$ \times $768, and around 50 times larger than the standard image size processed by image classification models, which is 224$ \times $224.}
\label{fig:gigapixel}
\end{figure*}

Many modern devices such as smartphones, drones, augmented reality headsets, vehicles and other Internet of Things (IoT) devices are equipped with high-quality cameras that can capture high-resolution images and videos. With the help of image stitching techniques, camera arrays \cite{thomson2021gigapixel, 7951481}, gigapixel acquisition robots \cite{sargent2010timelapse} and whole-slide scanners \cite{farahani2015whole}, capture resolutions can be increased to billions of pixels (commonly referred to as \textit{gigapixels}), such as the image depicted in Figure \ref{fig:gigapixel}. One could attempt to define \textit{high-resolution} based on the capabilities of human visual system. However, many deep learning tasks rely on data captured by equipment which behaves very differently compared to the human eye, such as microscopes, satellite imagery and infrared cameras. Furthermore, utilizing more detail than the eye can sense is beneficial in many deep learning tasks, such as in the applications discussed in Section \ref{sec:applications}. The amount of detail that can be captured and is useful if processed varies greatly from task to task. Therefore, the definition of high-resolution is \textit{task-dependent}. For instance, in image classification and computed tomography (CT) scan processing, a resolution of 512$ \times $512 pixels is considered to be high \cite{Chen2020, dosovitskiy2021an}. In visual crowd counting, datasets with High-Definition (HD) resolutions or higher are common \cite{gao2020cnn}, and whole-slide images (WSIs) in histopathology, which is the study of diseases of the tissues, or remote sensing data, which are captured by aircrafts or satellites, can easily reach gigapixel resolutions \cite{vanderLaak2021, van2018you}. 

Moreover, with the constant advancement of hardware and methodologies, what deep learning literature considers high-resolution has shifted over time. For instance, in the late 1990s, processing the 32$\times$32-pixel MNIST images with neural networks was an accomplishment \cite{726791}, whereas in early 2010s, the 256$\times$256-pixel images in ImageNet were considered high-resolution \cite{10.5555/2999134.2999257}. This trend can also be seen in the consistent increase of the average resolution of images in popular deep learning datasets, such as crowd counting \cite{gao2020cnn} and anomaly detection \cite{9271895} datasets. Therefore, the definition of high-resolution is also \textit{period-dependent}. Based on the task- and period-dependence properties, it is clear that the term ``high-resolution'' is technical, not fundamental or universal. Therefore, instead of trying to derive such a definition, we shift our focus to resolutions that create technical challenges in deep learning at the time of this writing.

\begin{figure*}
\begin{center}
\begin{tabular}{ c c c c }
\includegraphics[width=0.23\textwidth]{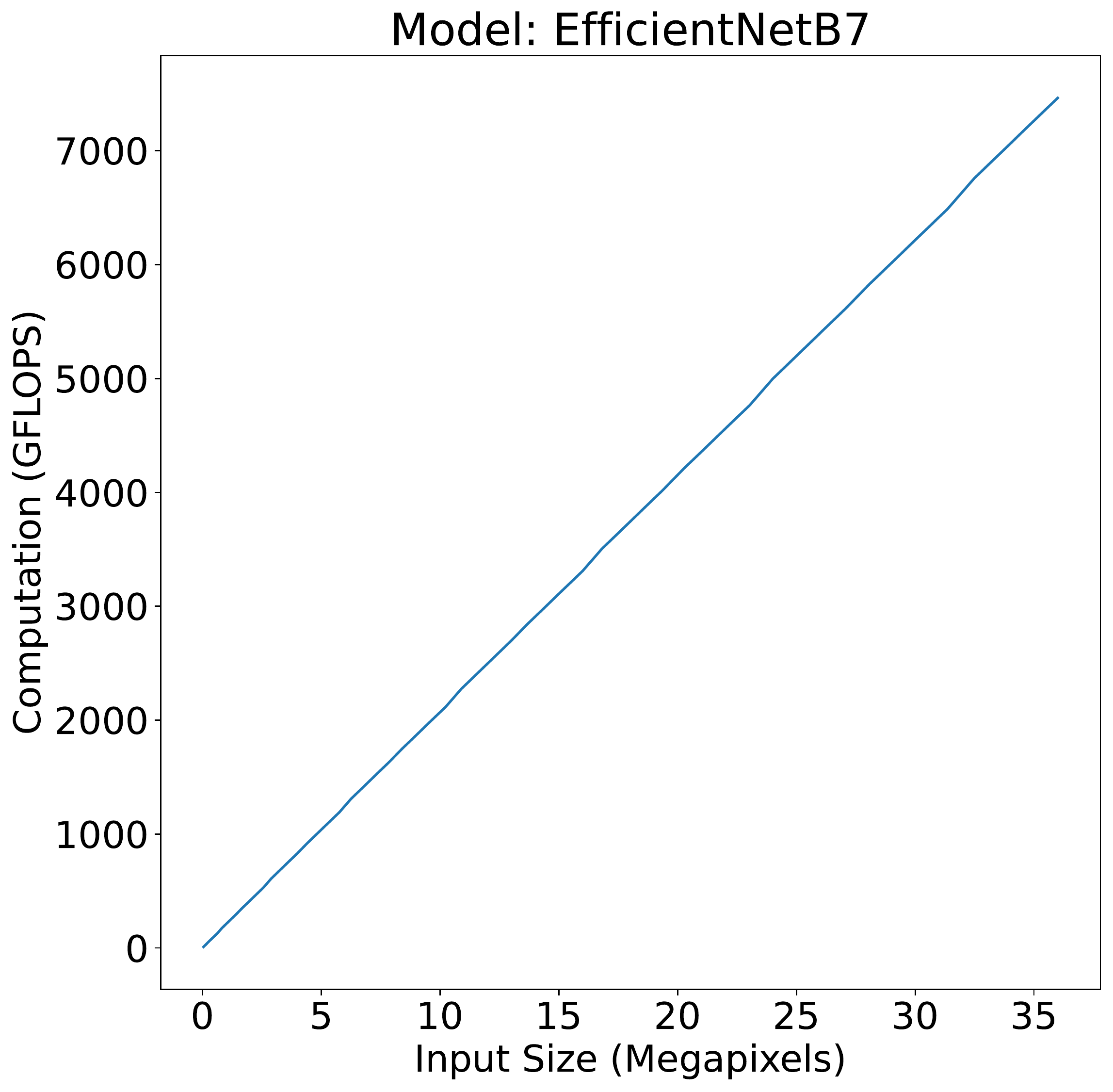} & \includegraphics[width=0.23\textwidth]{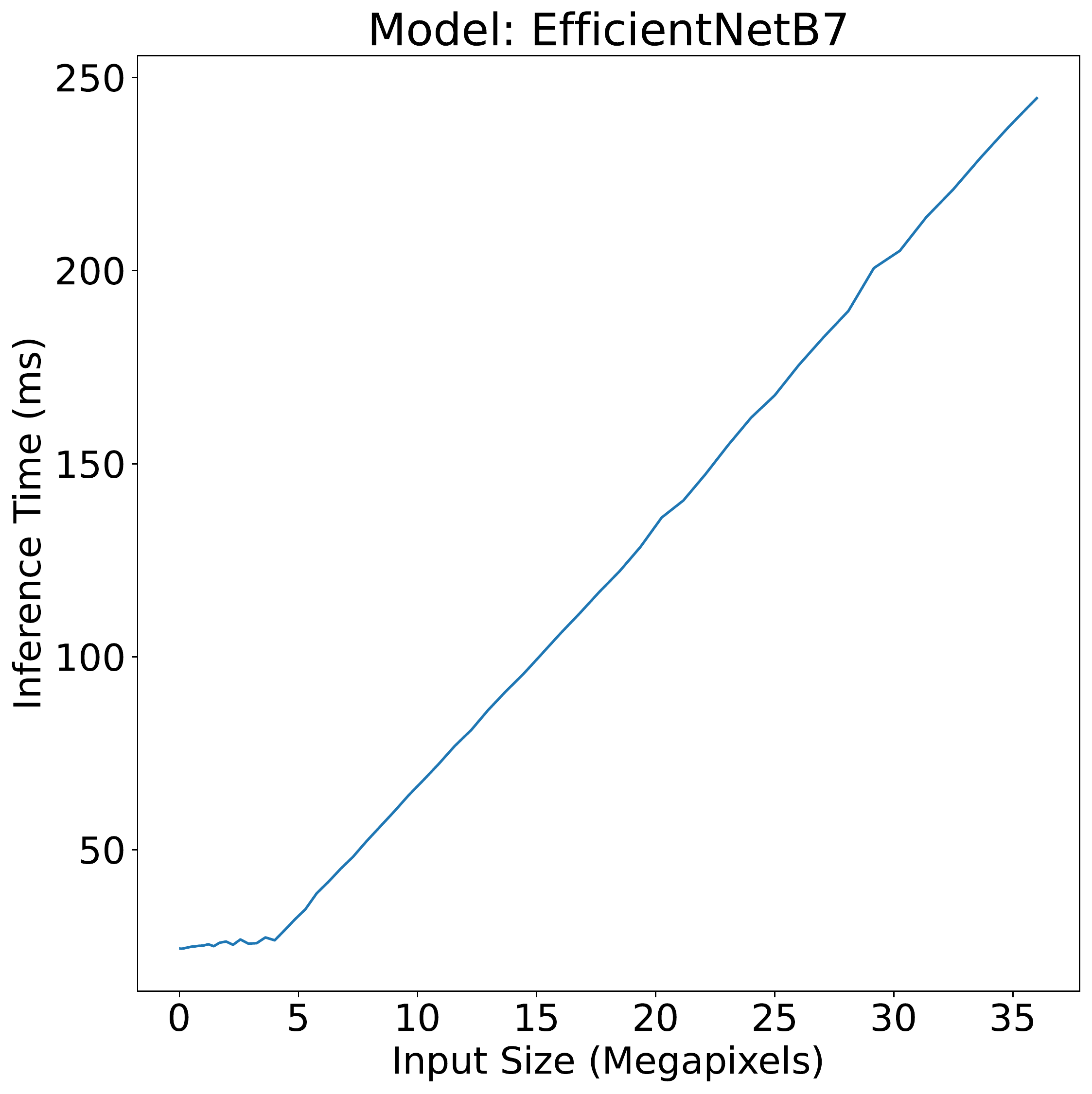} &
\includegraphics[width=0.23\textwidth]{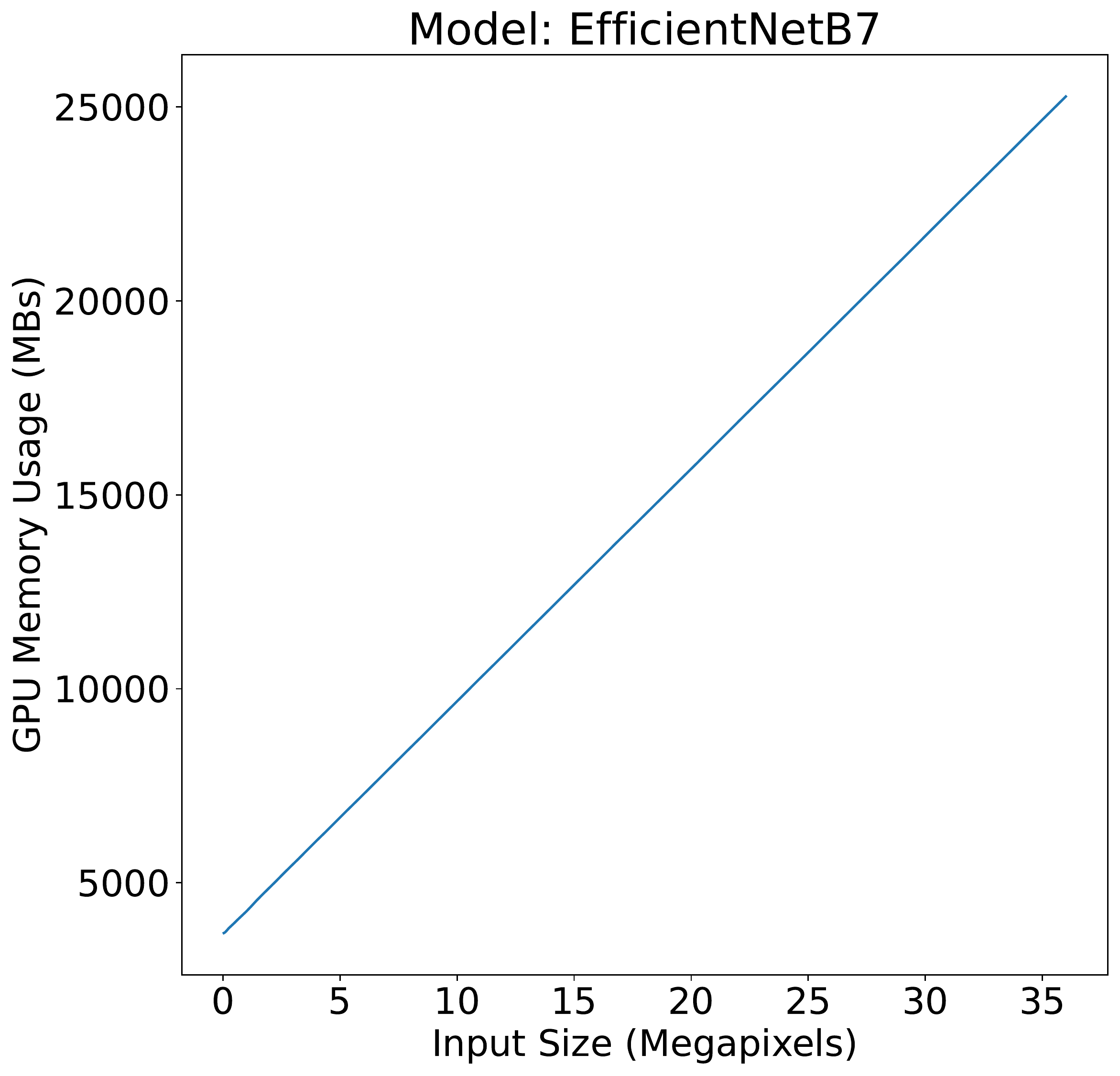} & \includegraphics[width=0.23\textwidth]{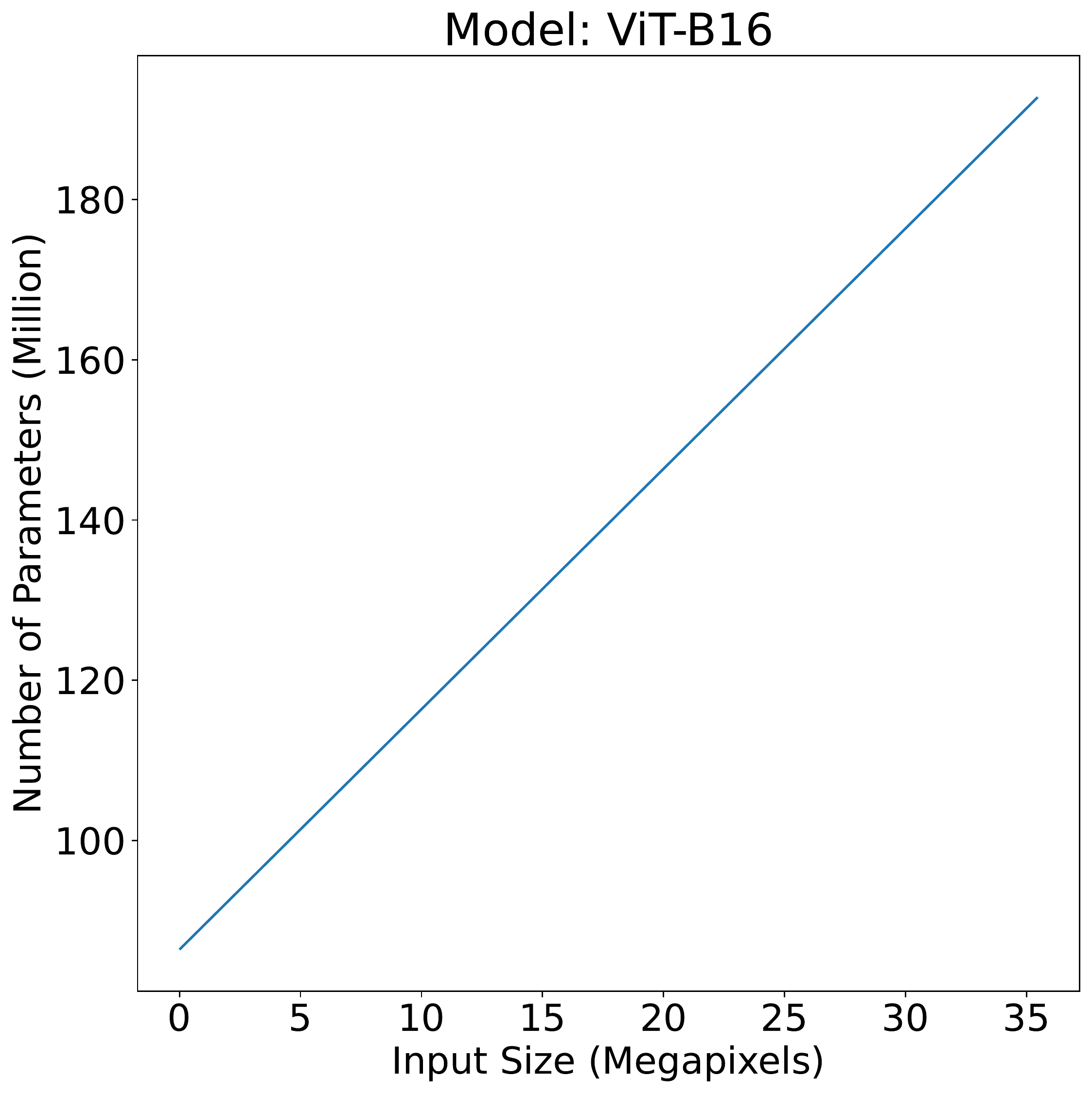}\\
 (a) & (b) & (c) & (d) \\
\end{tabular}
\end{center}
\caption{As the resolution of the input image increases, so does (a) the amount of computation, (b) inference time, and (c) GPU memory usage in the EfficientNet-B7 \cite{tan2019efficientnet}; and (d) the number of parameters in the ViT-B16 \cite{dosovitskiy2021an} architecture. The last layer of EfficientNet-B7 was removed to form a fully-convolutional feature extractor. Since accuracy is not considered in these figures, there is no need to use real images, thus randomly generated images are given to the models as input. All experiments were conducted on an Nvidia A6000 GPU.}
\label{fig:measurements}
\end{figure*}

Using high-resolution images and videos directly as inputs to deep learning models creates challenges during both training and inference phases. With the exception of fully-convolutional networks (FCNs), the number of parameters in deep learning models typically increases with larger input sizes. Moreover, the amount of computation, which is commonly measured in terms of floating point operations (FLOPs), and therefore inference/training time, as well as GPU memory consumption increase with higher-resolution inputs, as shown in Figure \ref{fig:measurements}. This issue is especially problematic in Vision Transformer (ViT) architectures, which use the self-attention mechanism, where the inference speed and number of parameters scale quadratically with input size \cite{dosovitskiy2021an, tay2020efficient}. These issues are exacerbated when the training or inference needs to be done on resource-constrained devices, such as smartphones, that have limited computational capabilities compared to high-end computing equipment, such as workstations or servers.

Even though methods such as \textit{model parallelism} can be used to split the model between multiple GPUs during both the training \cite{shoeybi2019megatron, openailarge} and inference \cite{du2020model} phases, and thus avoid memory and latency issues, these methods require a large amount of resources, such as a large number of GPUs and servers, which can incur high costs, especially when working with extreme resolutions such as gigapixel images. Furthermore, in many applications, such as self-driving cars and drone image processing, there is a limit for the hardware that can be mounted, and offloading the computation to external servers is not always possible because of unreliability of the network connection due to movement and the time-critical nature of the application. Therefore, the most common approach for deep learning training and inference is to load the full model on each single GPU instance. Multi-GPU setups are instead typically used to speed up the training by increasing the overall batch size, to test multiple sets of hyper-parameters in parallel or to distribute the inference load. Consequently, in many cases, there is an effective maximum resolution that can be processed by deep learning models. As an example, the maximum resolution for inference using SASNet \cite{song2021choose}, which is the state-of-the-art model for crowd counting on the Shanghai Tech dataset \cite{zhang2016single} at the time of this writing, is around 1024$ \times $768 (less than HD) on Nvidia 2080 Ti GPUs which have 11 GBs of video memory.

\begin{figure}[htbp]
\centerline{\includegraphics[width=0.48\textwidth]{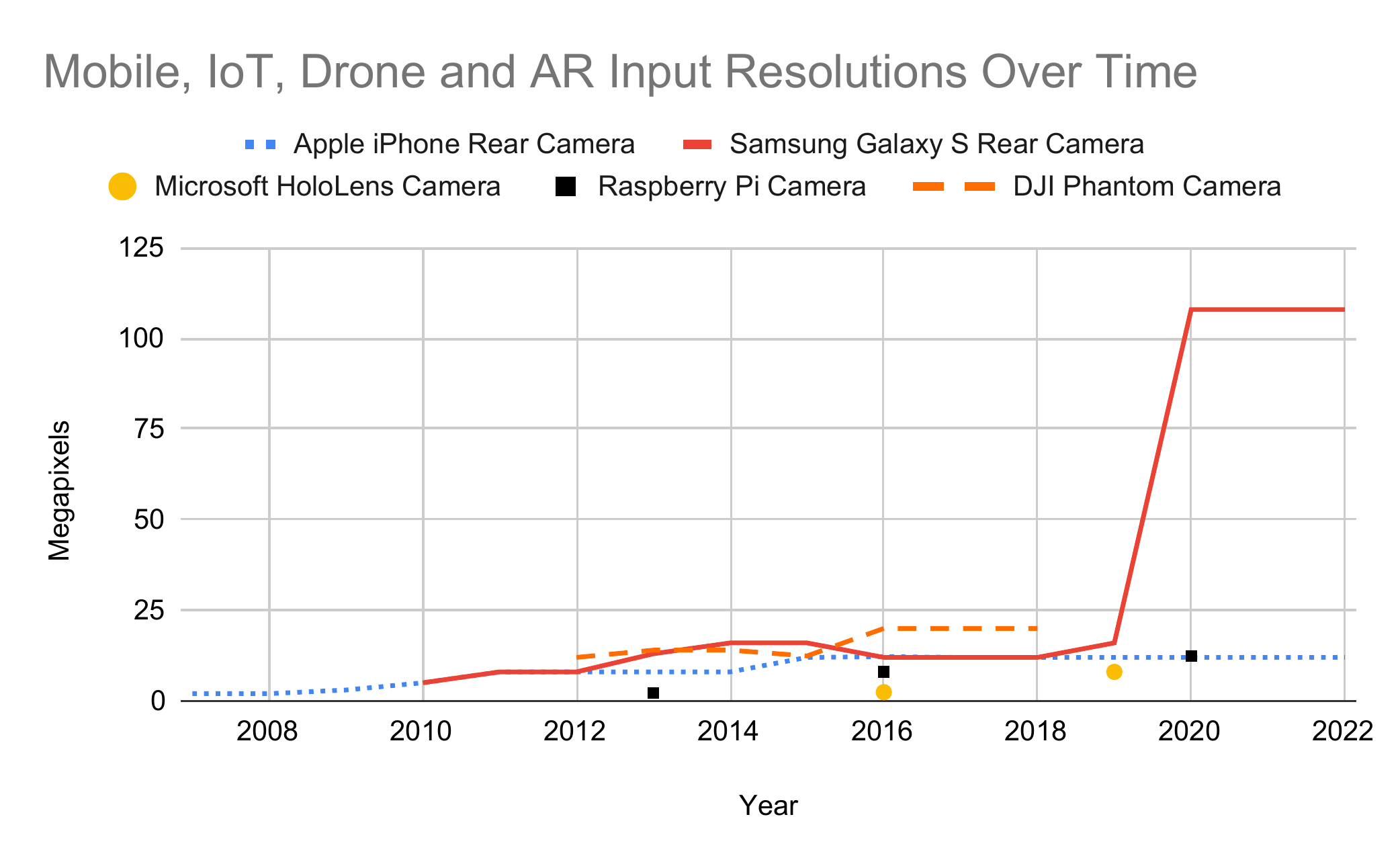}}
\caption{Trend of the maximum resolutions captured by smartphones (Apple iPhone and Samsung Galaxy S), drones (DJI Phantom), augmented reality headsets (Microsoft HoloLens) and IoT devices (Raspberry Pi) over time. Details and data sources are available in appendix \ref{sec:data_source}.}
\label{fig:trend}
\end{figure}

Although newer generations of GPUs are getting faster and have more memory available, the resolution of images and videos captured by devices is also increasing. Figure \ref{fig:trend} shows this trend across recent years for multiple types of devices. Therefore, the aforementioned issues will likely persist even with advances in computation hardware technology.  Furthermore, current imaging technologies are nowhere near the physical limits of image resolutions, which is estimated to be in petapixels \cite{10.1117/12.2014274}.

Whether or not capturing and processing a higher resolution leads to improvements depends on the particular problem at hand. For instance, in image classification, it is unlikely that increasing the resolution for images of objects or animals to gigapixels would reveal more beneficial details and improve the accuracy. On the other hand, if the goal is to count the total number of people in scenes such as the one presented in Figure \ref{fig:gigapixel}, using an HD resolution instead of gigapixels would mean that several people could be represented by a single pixel, which significantly increases the error. Similarly, it has been shown that using higher resolutions in histopathology can lead to better results \cite{lu2020capturing}.

Assuming there is an effective maximum resolution for a particular problem due to hardware limitations or latency requirements, there are two simple baseline approaches for processing the original captured inputs which are commonly used in deep learning literature \cite{recasens2018learning, 9127795, 10.3389/fmed.2019.00264}. The popularity of these baselines can be attributed to the simplicity of their implementation. The first one is to resize (downsample) the original input to the desired resolution, however, this will lead to a lower accuracy if any important details for the problem at hand are lost. This approach is called \textit{uniform downsampling (UD)} since the quality is reduced uniformly throughout the image. The second approach is to cut up the original input into smaller patches that each have a maximum resolution, process the patches independently, and aggregate the results, for instance, by summing them up for regression problems and majority voting for classification problems. We call this approach \textit{cutting into patches (CIP)}. There are two issues with this approach. First, many deep learning models rely on global features which will be lost since features extracted from each patch will not be shared with other patches, leading to decreased accuracy. For instance, crowd counting methods typically heavily rely on global information such as perspective or illumination \cite{gao2020cnn}, and in object detection, objects near the boundaries may be split between multiple patches. Secondly, since multiple passes of inference are performed, that is, one pass for each patch, inference will take much longer. This issue is worse when patches overlap.

To highlight these issues, we test the two baseline approaches (UD and CIP) on the Shanghai Tech Part B dataset \cite{zhang2016single} for crowd counting, which contains images of size 1024$ \times $768 pixels. We reduce the original image size by factors of 4 and 16 and measure the mean absolute error (MAE) for both baselines. To test UD, we take a SASNet model \cite{song2021choose} pre-trained on the Shanghai Tech Part B dataset \cite{zhang2016single} with input size of 1024$ \times $768, and fine-tune it for the target input size using the AdamW optimizer \cite{DBLP:conf/iclr/LoshchilovH19} with a learning rate of $ 10^{-5} $ and weight decay of $ 10^{-4} $. Note that the original SASNet paper uses the Adam optimizer \cite{DBLP:journals/corr/KingmaB14} with a learning rate of $ 10^{-5} $. We train the model for 100 epochs with batch size of 12 per GPU instance using 3$ \times$Nvidia A6000 GPUs. We empirically found that fine-tuning does not improve the accuracy of cutting into patches, therefore, we cut the original image into 4 and 16 patches, and obtain the count for each patch using the pre-trained SASNet mentioned above, then aggregate the results by summing up the predicted count for each patch.

The results of these experiments are shown in Table \ref{tab:baselines}. It can be observed that uniform downsampling significantly increases the error compared to processing the original input size. Keep in mind that even though the increase in error is not as drastic with cutting into patches, the inference time of this approach is increased by the same factor (i.e., 4 and 16) since we assumed we are using the effective maximum resolution possible for our hardware, and thus patches cannot be processed in parallel as the entire hardware is required to process a single patch.

\begin{table}[htbp]
\caption{Performance of baseline approaches on the Shanghai Tech Part B dataset.}
\begin{center}
\resizebox{\linewidth}{!}{
\begin{tabular}{ l | c c c} 
\hline
Input Size & Original MAE & UD$^{*}$ MAE & CIP$^{\dagger}$ MAE\\
\hline
\hline
1024$ \times $768 (original) & 6.31 & - & -\\
512$ \times $384 (reduced 4$ \times $) & - & 9.01 (\textcolor{red}{+43\%}) & 6.40 (\textcolor{red}{+1\%})\\
256$ \times $192 (reduced 16$ \times $) & - & 16.06 (\textcolor{red}{+155\%}) & 6.67 (\textcolor{red}{+6\%})\\
\hline
\multicolumn{4}{l}{$^{*}$Uniform Downsampling}\\
\multicolumn{4}{l}{$^{\dagger}$Cutting into Patches}\\
\end{tabular}
}
\end{center}
\label{tab:baselines}
\end{table}

Since these baseline approaches are far from ideal, in recent years, several alternative methods have been proposed in the literature in order to improve accuracy and speed while complying with the maximum resolution limitation caused either by memory limitations or speed requirements. The goal of this survey is to summarize and categorize these contributions. To the best of our knowledge, no other survey on the topic of high-resolution deep learning exists. However, there are some surveys that include aspects relevant to this topic. A survey on methods for reducing the computational complexity of Transformer architectures is provided in \cite{tay2020efficient}, which discusses the issues related to the quadratic time and memory complexity of self-attention and analyzes various aspects of efficiency including memory footprint and computational cost. While reducing the computational complexity of Transformer models can contribute to efficient processing of high-resolution inputs, in this survey, we only include Vision Transformer methods that explicitly focus on high-resolution images. Some application-specific surveys include high-resolution datasets and methods that operate on such data. For instance, a survey on deep learning for histopathology, which mentions challenges with processing the giga-resolution of WSIs, is provided in \cite{SRINIDHI2021101813}; a survey of methods that achieve greater spatial resolution in computed tomography (CT) is provided in \cite{SCHUIJF2022388}, which highlights improved diagnostic accuracy with ultra high-resolution CT, and briefly discusses deep learning methods for noise reduction and reconstruction; a survey on crowd counting where many of the available datasets are high-resolution is provided in \cite{gao2020cnn}; a survey on deep learning methods for land cover classification and object detection in high-resolution remote sensing imagery is provided in \cite{rs12030417}; and a survey on deep learning-based change detection in high-resolution remote sensing images is provided in \cite{rs14071552}.

\begin{figure}[htbp]
\centerline{\includegraphics[width=0.48\textwidth]{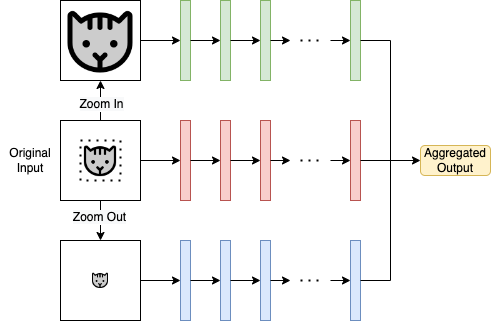}}
\caption{Schematic illustration of a multi-column architecture. If the original input to the DNN is a patch taken from a larger image, such as in \cite{9346018}, in addition to zooming in, it is also possible to zoom out.}
\label{fig:multi_column}
\end{figure}

It is important to mention that some methods operate on high-resolution inputs, yet do not make any effort to address the aforementioned challenges. For instance, \textit{multi-column} (also known as \textit{multi-scale}) networks \cite{gao2020cnn} incorporate multiple columns of layers in their architecture, where each column is responsible for processing a specific scale as shown in Figure \ref{fig:multi_column}. However, since the columns process the same resolution as the original input, most of these methods in fact require even more memory and computation compared to the case where only the original scale is processed. The primary goal of these methods is instead to increase the accuracy by taking into account the scale variances that occur in high-resolution images, although there are some multi-scale methods that improve both accuracy and efficiency \cite{10.1007/978-3-319-46493-0_22, 9052469, Zhao_2018_ECCV}. Therefore, these methods do not fall within the scope of this survey, unless they explicitly address the efficiency aspect for high-resolution inputs. ZoomCount \cite{9025247}, Locality-Aware Crowd Counting \cite{9346018}, RAZ-Net \cite{Liu_2019_CVPR} and Learn to Scale \cite{Xu_2019_ICCV} are all examples of multi-scale methods in crowd counting, and DMMN \cite{HO2021101866} and KGZNet \cite{8982943} in medical image processing.

The primary purpose of this survey is to collect and describe methods that exist in deep learning literature, which can be used in situations where the high resolution of input images and videos create the aforementioned technical challenges regarding memory, computation and time. The rest of this paper is organized as follows: Section \ref{sec:applications} lists applications where high-resolution images and videos are processed using deep learning. Section \ref{sec:methods} categorizes efficient methods for high-resolution deep learning into five general categories and provides several examples for each category. This section also briefly discusses alternative approaches for solving the memory and processing time issues caused by high-resolution inputs. Section \ref{sec:datasets} lists existing high-resolution datasets for various deep learning problems and provides details for each of them. Section \ref{sec:discussion} discusses the advantages and disadvantages of using efficient high-resolution methods belonging to different categories and provides recommendations about which method to use in different situations. Finally, Section \ref{sec:conclusion} concludes the paper by summarizing the current state and trends in high-resolution deep learning as well as suggestions for future research. The code for experiments conducted in this survey is available at \url{https://gitlab.au.dk/maleci/high-resolution-deep-learning}.

\section{Applications of High-Resolution Deep Learning}
\label{sec:applications}
In this section, we list some real-world applications where high-resolution images are processed with deep learning. Most of these methods do not focus on the efficiency angle, however, some of the methods address issues encountered with high-resolution images. For instance, \cite{MADEC2019225} mentions that ``it was not possible to train the model with the original 6000$ \times $4000 pixels images because of GPU memory limitation'' and \cite{xu2018road}, which uses the cutting into patches approach, states that ``a raw remote image has millions of pixels and is difficult to process directly''.

\subsection{Medical and Biomedical Image Analysis}
Multi-gigapixel whole-slide pathology images can be processed with deep learning in order to detect breast cancer \cite{liu2017detecting}, skin cancer \cite{Wang318, pmlr-v121-xie20a}, prostate cancer \cite{pmlr-v121-xie20a}, lung cancer \cite{pmlr-v121-xie20a}, cervical cancer \cite{Cheng2021} and cancer in the digestive tract \cite{8983226}. Some methods are even able to detect the cancer subtypes \cite{pmlr-v121-xie20a} or detect the spread of cancer to lymph nodes (metastasis) \cite{8604098}. Semantic segmentation of such images can be useful in neuropathology \cite{Lai_2021_ICCV}, which is the study of diseases of the nervous system, and identifying tissue components such as tumor, muscle, and debris in medical images \cite{Javed_2019_ICCV}.

Moreover, the processing of high-resolution computed tomography (CT) scans with deep learning is becoming more prevalent. The studies in \cite{Yang2020} and \cite{Chen2020} detect COVID-19 in high-resolution CT scans of the lung, and the study in \cite{Akagi2019} uses deep learning to improve the quality of captured ultra-high-resolution CT scans. In addition, the study in \cite{10.1093/bioinformatics/btaa1094} performs semantic segmentation on high-resolution electron microscopy images from hearts and brains of mice, which is useful for fundamental biomedical research. Additionally, high-resolution deep learning can be used for reconstruction of CT images and reduction of image noise, which has been shown to obtain results similar to other conventional methods with clinically feasible speed \cite{FUKUSHIMA2022110294, MCLEAVY2021407}.

Even though medical image analysis methods primarily focus on improving the accuracy of particular tasks, inference speed can be crucial in some applications, for instance, speed might be a requirement in clinical practice \cite{8604098}. Furthermore, real-time augmented reality under microscopes can provide suitable human–computer interaction for AI-assisted slide screening \cite{Cheng2021}. Finally, there might be situations where the speed for processing a single input is acceptable, however, the sheer number of input data is so high that inputs collectively cannot be processed within a deadline. For instance, 55,000 high-resolution images are taken during the examination of a single patient using wireless capsule endoscopy, where a tiny wireless camera is swallowed to take pictures of the digestive tract, which can be used to detect lesions and inflammation \cite{xing2020zoom}.

\subsection{Remote Sensing}
Processing high-resolution aerial and satellite imagery with deep learning has various applications \cite{ball2017comprehensive}, such as detecting buildings \cite{7326158}, which is useful for urban planning and monitoring; detecting airplanes \cite{alganci2020comparative}, which can be used for defense and military applications as well as airport surveillance; extracting road networks \cite{xu2018road}, which has applications in automated road navigation with unmanned vehicles, urban planning and real-time updating of geospatial databases; detecting areas in a forest that are damaged due to natural disasters such as storms \cite{hamdi2019forest}; identifying weed plants, which can be used for targeted spraying of pesticides in agricultural fields; semantic segmentation of satellite data which can help with crop monitoring, natural resource management and digital mapping \cite{9759447}; and remote sensing image captioning which is useful for applications such as image retrieval and military intelligence generation \cite{9400386}. Moreover, significant accuracy improvements can be obtained by taking low-resolution weather data as input and interpolating high-resolution data using super-resolution \cite{8588749}. The motivation behind this approach is that high-resolution data are only available with a few days delay, and this method can be used to more accurately process low-resolution but up-to-date data.

\subsection{Surveillance}
\begin{figure}[htbp]
\centerline{\includegraphics[width=0.3\textwidth]{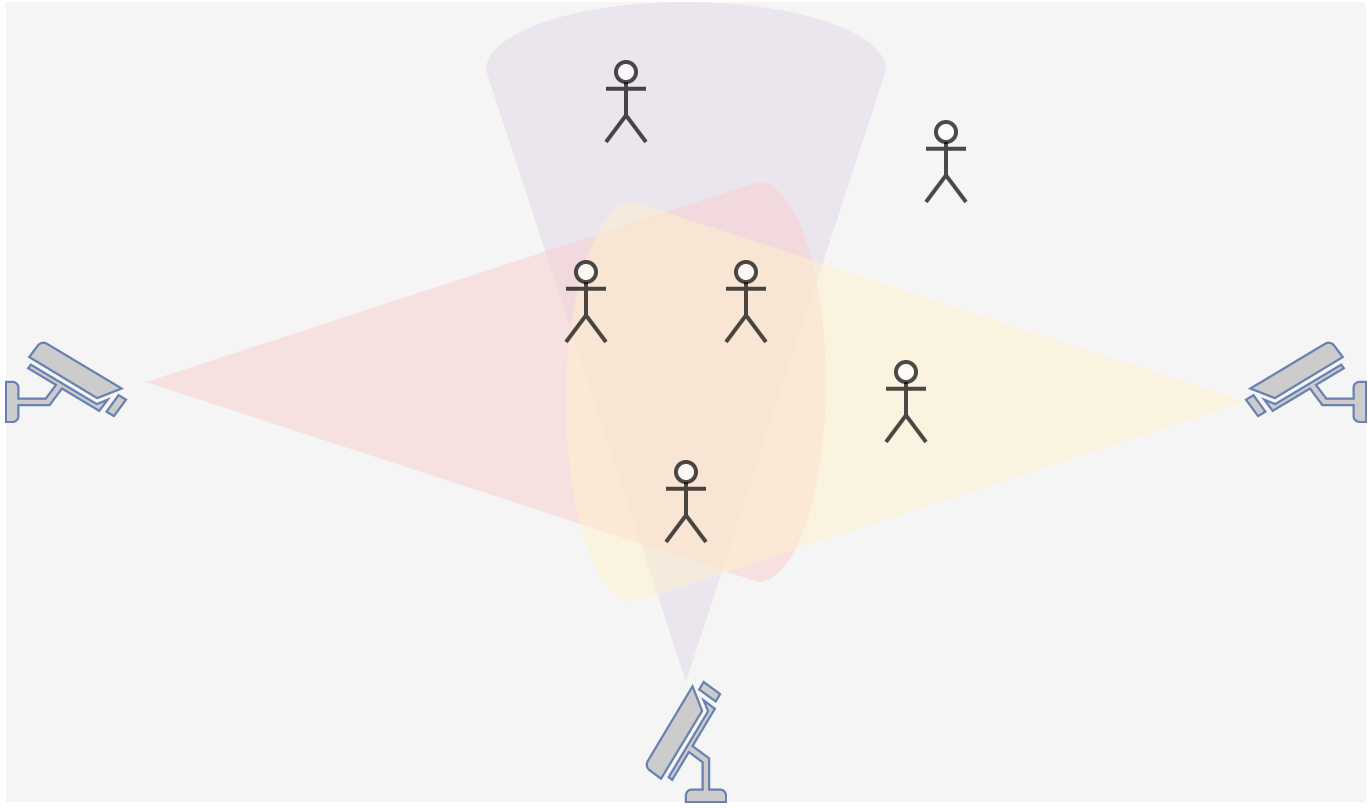}}
\caption{Overlap in the field of view for multi-camera setups, which can result in duplicates in tasks such as crowd counting.}
\label{fig:multi_view}
\end{figure}

Capturing and processing gigapixel images for surveillance is becoming increasingly widespread, and such images can be processed with deep learning for searching and identifying people \cite{10.1007/978-3-030-64559-5_30, Specker_2022_WACV} as well as detecting pedestrians \cite{9535550, LI2022482} which can be used for human behavior analysis and intelligent video surveillance such as enforcing social distancing restrictions during a pandemic \cite{proxemics2021,ahmed2021socialDistance}. It should be noted that capturing gigapixel images for surveillance has several advantages over capturing lower resolutions with multiple cameras at different locations of the scene. First, cameras in a multi-camera setup typically have some overlap in their fields of view to avoid blindspots. This may result in errors for many applications, such as crowd counting, due to duplicates, as shown in Figure \ref{fig:multi_view}. Reducing this error is not an easy task, since it requires information about the geometry of the scene and the use of re-identification methods for identifying and deduplicating people in multiple views of the same scene. Secondly, tracking the trajectory of people, vehicles and other moving objects is difficult with multiple cameras, since it also requires identifying them in multiple views of the scene. Finally, in many deep learning applications such as crowd counting, incorporating global information from the entire scene such as illumination and perspective improves the accuracy of the task \cite{gao2020cnn}. Note that images captured from drastically different locations and perspectives, such as the ones in in Figure \ref{fig:multi_view}, cannot be stitched together to form a single image.

\subsection{Other Applications}
High-resolution deep learning can be beneficial in many other applications and various domains of science. For instance, the study in \cite{MADEC2019225} estimates the density of wheat ears, which are the grain-bearing parts of the plant, from high-resolution images taken from grain fields, which aids plant breeders in optimizing their yield; and the study in \cite{Horwath2020} introduces a deep learning method for segmentation of high-resolution electron microscopy images, which has applications in material science such as understanding the degradation process of industrial catalysts. \cite{Lin_2021_CVPR} proposes a method for real-time high-resolution background replacement, which is useful in video calls and conferencing.

\section{Methods for Efficient Processing of High-Resolution Inputs with Deep Learning}
\label{sec:methods}

\subsection{Non-Uniform Downsampling}
\label{sec:non_uniform_downsampling}
\begin{figure}[htbp]
\centerline{\includegraphics[width=0.48\textwidth]{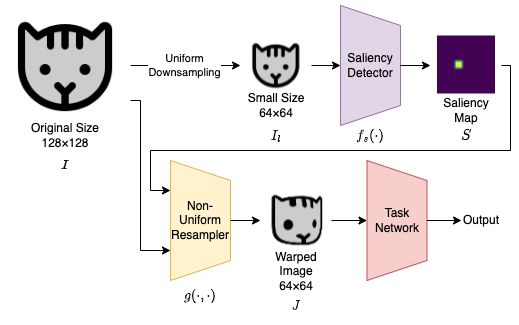}}
\caption{Schematic illustration of the non-uniform downsampling approach. The saliency detector detects the cat's right eye as a salient area, therefore, the non-uniform resampler samples more pixels from that area.}
\label{fig:non_uniform_downsampling}
\end{figure}

\textit{Non-uniform downsampling (NUD)} is based on the idea that for any deep learning task, some locations of an input image are more important than others. For instance, in gaze estimation, where the goal is to detect where a person is looking given an image including the person's face, the image locations depicting the person's eyes are much more important than other parts of the image. Therefore, when reducing the resolution of the image, it might be beneficial to sample more pixels from salient areas and less pixels from non-salient locations, resulting in a warped and distorted image. This operation requires salient areas to be determined before introducing the downsampled image to the task DNN. Therefore, a small saliency detection network is utilized in order to obtain this saliency map. Figure \ref{fig:non_uniform_downsampling} provides a schematic illustration of the non-uniform downsampling approach. Note that non-uniform downsampling is a broad process that encompasses any method that downsamples the input image in any manner other than uniform. \cite{recasens2018learning} further subdivides non-uniform downsampling into three categories: attention mechanisms, saliency-based methods and adaptive image sampling methods. However, as the authors point out, there is a lot of overlap between these categories and it is difficult to draw a clear border between them.

Formally, the saliency map $ S $ can be obtained by applying saliency detection network $ f_s(\cdot) $ on a uniformly downsampled image $ I_l $, that is, $ S = f_s(I_l) $. The input to the saliency detection network is downsampled in order to keep the overhead of the saliency detection process low. The non-uniformly downsampled image $ J $ can the be obtained based on $ J = g(I, S) $, where $ g(\cdot) $ is the non-uniform resampler and $ I $ is the original image. Essentially, the resampler should compute a mapping $ J(x, y) = I(u_c(x, y), v_c(x, y)) $ from the original image to the downsampled one. Functions $ u_c(\cdot) $ and $ v_c(\cdot) $ need to map pixels proportionally to the weight assigned to them in the saliency map. Assuming the saliency map is normalized and $ \forall x, y: 0 \leq u_c(x, y) \leq 1 $ and $ \forall x, y: 0 \leq v_c(x, y) \leq 1 $, this problem can be written as
\begin{equation}
\int_0^{u_c(x, y)} \int_0^{v_c(x, y)} S(x', y') dx' dy' = xy.
\label{eq:nud_integral}
\end{equation}

However, methods for determining this transformation based on Eq. \ref{eq:nud_integral} are not efficient \cite{recasens2018learning}. An alternative approach is to presume each pixel $ (x', y') $ is pulling all other pixels with a force proportional to its saliency $ S(x', y') $, which can be formulated as
\begin{eqnarray}
u_c(x, y) = \frac{\sum_{x', y'} S(x', y') k((x, y), (x', y')) x'}{\sum_{x', y'} S(x', y') k((x, y), (x', y'))},
\label{eq:nud_conv_x}\\
v_c(x, y) = \frac{\sum_{x', y'} S(x', y') k((x, y), (x', y')) y'}{\sum_{x', y'} S(x', y') k((x, y), (x', y'))},
\label{eq:nud_conv_y}
\end{eqnarray}
where $ k((x, y), (x', y')) $ is a distance kernel, for instance, the Gaussian kernel. Using this formulation, salient areas will be sampled more since they attract more pixels. Moreover, based on this formulation, $ u_c (\cdot) $ and $ v_c (\cdot) $ can be computed with simple convolutions. Therefore, this operation can be easily plugged into neural network architectures as a layer, and has the added benefit of preserving the differentiability which is a requirement for training neural networks with the backpropagation algorithm. The overall result is that the entire module including the saliency detection network and the task network can be trained end-to-end. The method in \cite{recasens2018learning} uses this approach to improve the performance of gaze estimation as well as fine-grained classification, which is the task of differentiating between hard-to-distinguish objects such as different species of animals.

The method in \cite{marin2019efficient} applies the idea of non-uniform downsampling to semantic segmentation. If the input image $ I = I_{ij} $ has a size $ H \times W $ and must be downsampled to size $ h \times w $, the first step is to generate ideal sampling tensors from ground truth (GT) labels based on
\begin{equation}
E(\phi) = \sum_{i, j} \| \phi_{ij} - b(u_{ij}) \|^2 + \lambda \sum_{|i - i'| + |j - j'| = 1} \| \phi_{ij} - \phi_{i'j'} \|^2,
\label{eq:segmentation_energy}
\end{equation}
where $ \phi \in [0,1]^{h \times w \times 2} $ is the sampling tensor to be determined, $ E(\phi) $ is the (energy) cost function to minimize, $ u  \in [0,1]^{h \times w \times 2} $ is the uniform downsampling tensor and $ b(u_{ij}) $ is the coordinates of the closest point to pixel $ u_{ij} $ on semantic boundaries in the GT labels. Eq. \ref{eq:segmentation_energy} corresponds to a least squares problem with convex constraints that can be efficiently solved using a set of sparse linear equations. The first term in Eq. \ref{eq:segmentation_energy} ensures the sampling locations are close to the semantic boundaries, and the second term ensures that the distortion is not excessive by forcing the transformations of adjacent pixels to be similar. Eq. \ref{eq:segmentation_energy} is also subject to covering constraints that ensure the sampled locations cover the whole image. The contribution of the second term is controlled by a parameter $ \lambda $ which is empirically set to 1. The next step is to train a neural network to generate sampling tensors from input images. The images are then downsampled based on the output of this neural network and introduced to the task network. Finally, the segmentation output is upsampled to remove distortions and match the original resolution.

Similarly, the method in \cite{jin2021learning} utilizes non-uniform downsampling for semantic segmentation. However, in contrast with the previous method, the saliency detector in this method is optimized based on the performance of semantic segmentation rather than external supervision signals. This method is similar to \cite{recasens2018learning}, however, applying a straightforward adaptation of \cite{recasens2018learning} to semantic segmentation does not perform well. To improve the performance, an \textit{edge loss} is added as a regularization term, which is calculated by using the mean squared error (MSE) between the deformation map $ d $ obtained by the saliency detector and target deformation map $ d_t $ calculated based on segmentation labels. To combat trivial solutions, the target deformation map has denser sampling around object boundaries and is formulated by $ d_t = f_{\text{edge}}(f_{\text{gauss}}(Y_{lr})) $, where $ Y_{lr} $ is the uniformly downsampled segmentation label, $ f_{\text{edge}} $ is an edge detection filter by convolution with a specific $ 3\times3 $ kernel, and $ f_{\text{gauss}} $ is Gaussian blur with $ \sigma = 1 $.

Since the distortions caused by the \textit{customized grids} defined in Eqs. \ref{eq:nud_conv_x} and \ref{eq:nud_conv_y} can be severe, the method in \cite{xing2020zoom} introduces \textit{structured grids} that can be combined with customized grids to obtain a more subtle spatial distortion effect for wireless capsule endoscopy (WCE) image classification. These structured grids ensure that pixels that were in the same row/column in the input image are also in the same row/column in the output image, and can be obtained by
\begin{eqnarray}
u(x) = \frac{\sum_{x'} S(x') k(x, x') x'}{\sum_{x'} S(x') k(x, x')},
\label{eq:structured_grid_u}\\
v(y) = \frac{\sum_{y'} S(y') k(y, y') x'}{\sum_{y'} S(y') k(y, y')},
\label{eq:structured_grid_v}
\end{eqnarray}
where $ S(x) = \max_y S(x, y) $ and $ S(y) = \max_x S(x, y) $. $ u(x) $ and $ v(y) $ are then copied and stacked to form $ u_s(x, y) = u(x) $ and $ v_s(x, y) = v(y) $. Finally, the combined deformation grids can be computed by
\begin{eqnarray}
u(x, y) = \lambda u_s(x, y) + (1 - \lambda) u_c(x, y),
\label{eq:combined_u}\\
v(x, y) = \lambda v_s(x, y) + (1 - \lambda) v_c(x, y),
\label{eq:combined_v}
\end{eqnarray}
where parameter $ \lambda $ is empirically set to 0.5.

Similarly, FOVEA \cite{thavamani2021fovea} discards custom grids and solely relies on structured grids for object detection in autonomous driving use cases. It also introduces \textit{anti-cropping regularization} to combat cropping which may result in missing objects, by using reflect padding on the saliency map. In \cite{recasens2018learning}, the saliency detector is trained end-to-end along with the task network, however, as mentioned, finding saliency maps in object detection is more difficult. Therefore FOVEA uses intermediate supervision to train the saliency detection network.

\begin{figure}[htbp]
\centerline{\includegraphics[width=0.4\textwidth]{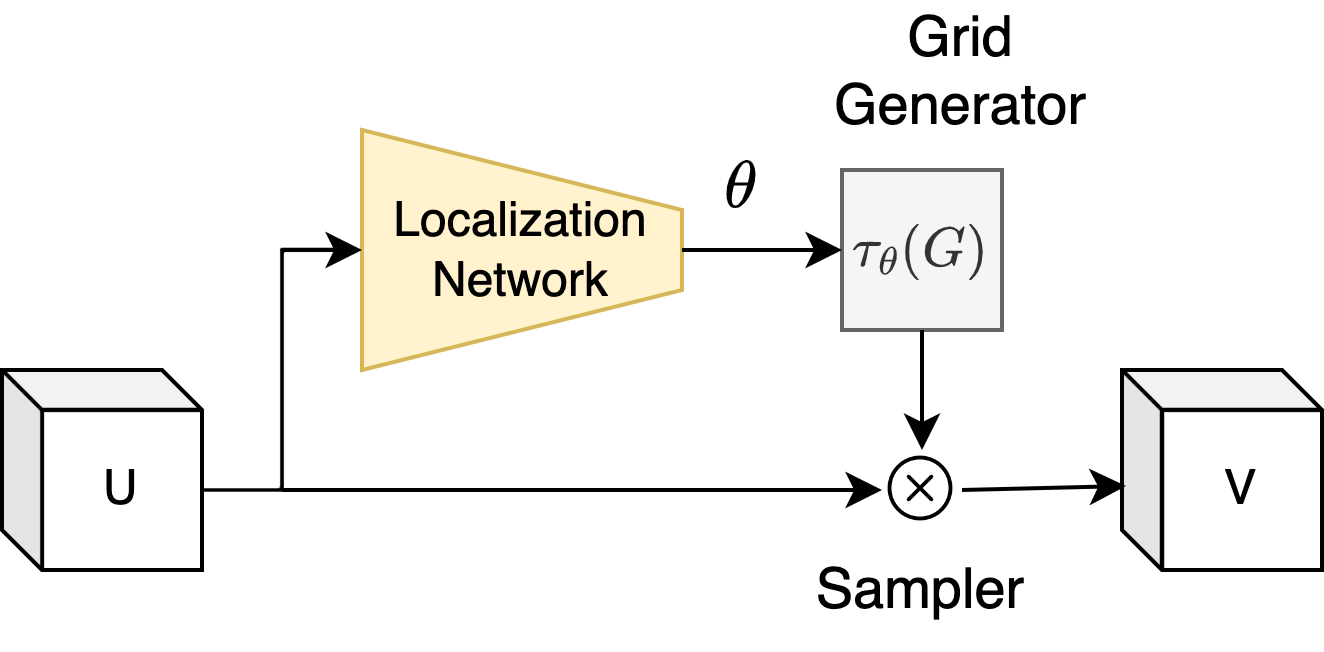}}
\caption{Architecture of the spatial transformer module \cite{jaderberg2015spatial}.}
\label{fig:st}
\end{figure}

Even though the primary goal of the \textit{spatial transformer} module in spatial transformer networks (STNs) \cite{jaderberg2015spatial} is to learn invariance to translation, scale, rotation and warping in order to improve performance, in the special case where the module is the first layer of the network, it can learn to crop the raw high-resolution input to a lower resolution and increase computational efficiency, thus it could be considered a form of NUD. Figure \ref{fig:st} shows the architecture of the spatial transformer module, where the localization network determines the parameters $ \theta $ for the transformation $ \tau_{\theta} $ from input features $ U $. $ \tau_{\theta}(\cdot) $ can be a 2D affine transformation, a more constrained transformation such as
\begin{equation}
A_{\theta} = \begin{bmatrix}
s & 0 & t_x\\
0 & s & t_y
\end{bmatrix},
\label{eq:constrained_transformation}
\end{equation}
which only allows cropping, translation and scaling, or a more general transformation such as plane projective transformation with 8 parameters, piecewise affine, thin plate spline \cite{10.1007/BFb0086566}, or any transformation as long as it is differentiable with respect to its parameters.

\begin{figure}[htbp]
\centerline{\includegraphics[width=0.48\textwidth]{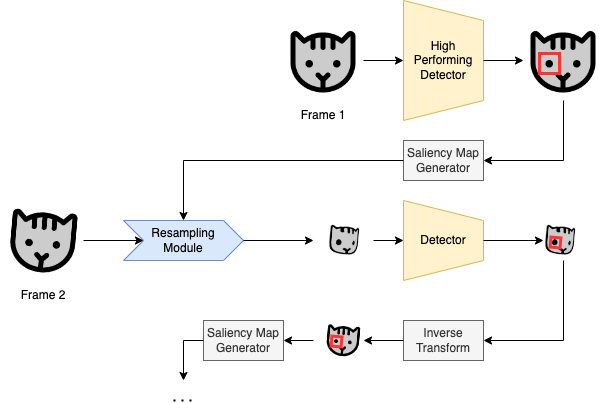}}
\caption{Overview of SALISA \cite{bejnordi2022salisa}. The second frame is slightly different from the first frame (in this case, slightly rotated clockwise), therefore, the detection result obtained from the first frame can be used to estimate the saliency of objects in the second frame.}
\label{fig:SALISA}
\end{figure}

SALISA \cite{bejnordi2022salisa} uses spatial transformer modules to perform non-uniform downsampling for object detection in high-resolution videos. In SALISA, the output of a video frame is used to determine the saliency map for the next frame. Figure \ref{fig:SALISA} shows this method, where the first frame is introduced to a high-performing detector without any downsampling. The detected objects are subsequently used to create a saliency map, which is then given to the resampling module. The resampling module contains a spatial transformer module with a thin plate spline transformation, where the localization network receives the saliency map as input. The downsampled image provided by the resampling module is then introduced to a lightweight detector. Since the lightweight detector detects objects in the warped image, the detected bounding boxes need to be transformed back into the original grid. Therefore an inverse transformation is applied before generating the saliency map. To prevent cascading errors, the method is reset to use the original high-resolution frame and high-performing detector every few frames.

\subsection{Selective Zooming and Skipping}
\label{sec:szs}

\textit{Selective zooming and skipping (SZS)} methods take a more efficient approach to cutting into patches by only zooming into regions of the input image that are important. The zoom level may differ across different patches, and some patches may be entirely skipped. \textit{Reinforced Auto-Zoom Net (RAZN)} \cite{dong2018reinforced} uses reinforcement learning to determine where to zoom in WSIs for the task of breast cancer segmentation. RAZN assumes the zoom-in action can be performed at most $ m $ times and the zooming rate is a constant $ r $. At each zoom level $ i $, there is a different segmentation network $ f_{\theta_i} $ and a different policy network $ g_{\theta_i} $. Initially, policy network $ g_{\theta_0} $ takes a cropped image $ x_0 \in \mathbb{R}^{H \times W \times 3} $ as input and determines whether to zoom-in or to break. If there is no need to zoom in, $ x_0 $ is given as input to segmentation network $ f_{\theta_0} $ which produces the output, otherwise, a higher-resolution image $ \hat{x}_0 \in \mathbb{R}^{rH \times rW \times 3} $ is sampled from the same area and will be cut into $ r^2 $ patches of size $ H \times W \times 3 $. Each patch is then given to policy network $ g_{\theta_1} $ and this process is recursively repeated until all policy networks break or the maximum zoom level is reached. RAZN achieves an improved performance over other state-of-the-art methods while reducing the inference time by a factor of $ \sim $2. Similarly, the methods in \cite{Gao_2018_CVPR} and \cite{Uzkent_2020_CVPR} use reinforcement learning for efficient object detection and aerial image classification, respectively.

Instead of reinforcement learning, the method in \cite{du2019zoom} uses a hierarchichal graph neural network to classify whether a mammogram (X-ray image of a breast) is normal/benign (contains a tumor that is not cancerous) or malignant (contains a tumor that is cancerous). At each zoom level $ i $, the graph $ G^i $ is defined by the adjacency matrix $ A^i \in \mathbb{R}^{N_i \times N_i}$ where there is an edge between each zoomed-in patch and its original image. The feature matrix of the graph is defined as $ X_i \in \mathbb{R}^{N_i \times D \times D} $, and the maximum zoom level is $ R $. The features on the nodes are zoomed-in regions of the input image, resized to $ D \times D $. A pre-trained CNN is used to extract feature vectors $ H_i \in \mathbb{R}^{N_i \times H} $ from $ X_i $. $ \text{GAT}_{\text{node}}(\cdot) $ is a graph attention network \cite{DBLP:conf/iclr/VelickovicCCRLB18} used to classify whether to zoom in for each node. Therefore, the output of the $ i $-th level in the hierarchical graph is
\begin{equation}
P_i = \begin{cases}
  1, & i = 1, \\
  \text{softmax}(\text{GAT}_{\text{node}}(A_i, H_i)), & 1 < i < R,
\end{cases}
\label{eq:hierarchical_graph}
\end{equation}
where $ P_i \in \mathbb{R}^{N_i \times 2} $ represents the decision to zoom or not for each node of the $ i $-th level. At the final zoom level $ R $, another graph attention network $ \text{GAT}_{\text{graph}}(\cdot) $ is used to perform the final classification for the entire mammogram based on $ \hat{Y} = \text{softmax}(\text{GAT}_{\text{graph}}(A_R, H_R)W) $, where $ W $ is a trainable weight matrix. The loss function contains both node losses and graph losses, with the zoom labels for nodes being obtained from lesion segmentation labels. This method achieves an accuracy comparable to the state-of-the-art, however, it is unclear how much it improves the inference speed.

GigaDet \cite{CHEN202214} achieves near real-time object detection in gigapixel videos. At the core of GigaDet is the \textit{Patch Generation Network (PGN)}. PGN takes a uniformly downsampled image as input and outputs a dense regression map which counts the number of objects that are completely contained within the corresponding area in the image, referred to as the \textit{patch candidate}. PGN is applied at different scales in order to obtain patch candidates of varying scales. The patch candidates selected by the PGN go through post-processing which includes non-maximum suppression (NMS), and are subsequently sorted based on their count. The top $ K $ patch candidates are then selected to be processed by the \textit{Decorated Detector (DecDet)} to detect objects. VGG \cite{simonyan2014very} and YOLO \cite{redmon2016you} are used for the PGN and DecDet networks, respectively. Given gigapixel videos, GigaDet is capable of running 5 FPS on a single Nvidia 2080 Ti GPU, which is $ 50 \times $ faster than Faster RCNN \cite{ren2015faster}, yet obtains a comparable performance in terms of average precision.

\begin{figure}[htbp]
\centerline{\includegraphics[width=0.4\textwidth]{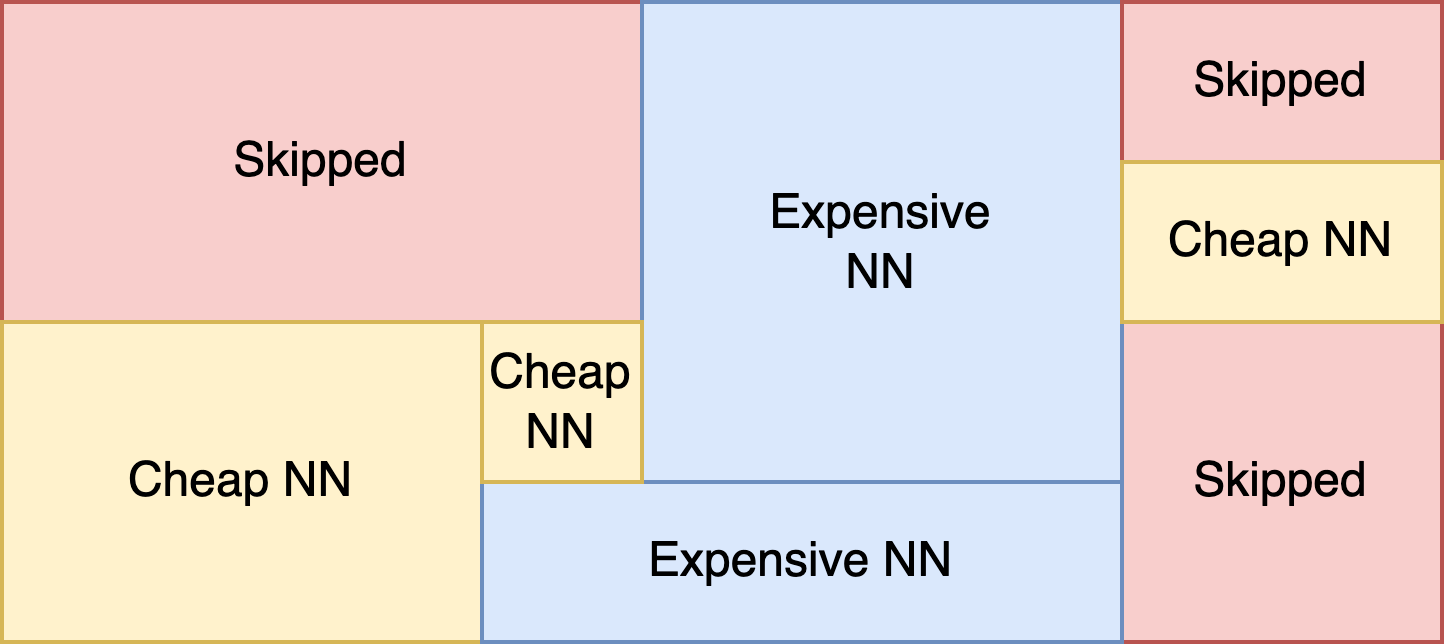}}
\caption{Partitioning in REMIX \cite{10.1145/3447993.3483274}. Some parts of the image are skipped, some processed by computationally cheap DNNs and some by computationally expensive DNNs.}
\label{fig:REMIX}
\end{figure}

REMIX \cite{10.1145/3447993.3483274} detects pedestrians in high-resolution videos within a latency budget given by the user. The input frame is partitioned into several blocks, where more salient blocks are processed using a computationally expensive but accurate network whereas less salient blocks are processed using a computationally cheap network or even skipped, as shown in Figure \ref{fig:REMIX}. REMIX uses historical frames to determine the object distribution, and determines the optimal partition using a dynamic programming algorithm that takes into account the given latency budget, the estimated object distribution, as well as the accuracy and speed of available neural networks for object detection. REMIX achieves up to $ 8.1 \times $ inference speedup with an accuracy comparable to the state-of-the-art methods. 

\subsection{Lightweight Scanner Networks}
\label{sec:lsn}

\textit{Lightweight scanner networks (LSNs)} are lightweight fully convolutional neural networks (FCNs) that efficiently scan the entire high-resolution input. To achieve a lightweight architecture, LSNs are typically designed and trained for very specific tasks. Moreover, as opposed to the cutting into patches approach, FCNs are inherently efficient in a sliding-window setting since they share the computation in overlapping regions \cite{sermanet2013overfeat}.

VGG-720p and VGG-1080p \cite{Tzelepi2020, TZELEPI2020107407} are LSNs capable of running in real-time on drones and provide heatmaps for input images of size 1280$ \times $720 and 1920$ \times $1080 pixels, respectively, that specify whether or not there are people, faces, or bicycles at each location in the input image. Both models take patches of size 32$ \times $32 or 64$ \times $64 pixels as input. The architectures of VGG-720p and VGG-1080, shown in Tables \ref{tab:vgg_720p} and \ref{tab:vgg_1080p}, respectively, contain only 5 convolutional layers with only 2 to 24 output channels. In contrast, the original VGG architectures have 11 to 19 layers with up to 512 output channels in some layers \cite{simonyan2014very}.

\begin{table}[htbp]
\caption{Architecture of VGG-720p.}
\begin{center}
\resizebox{\linewidth}{!}{
\begin{tabular}{ l c c c c c } 
\hline

Layer & Kernel & Stride & Pad$ ^{\dagger} $ (X/Y)$ ^* $ & Max Pool (X/Y) & Channels\\

\hline
\hline

conv1\_1 & 3$ \times$3 & 1/1 & 1/1 & - / - & 16\\
conv1\_2 & 3$ \times$3 & 1/1 & 1/1 & \checkmark / - & 16\\
conv2\_1 & 3$ \times$3 & 1/1 & 1/1 & - / - & 24\\
conv2\_2 & 3$ \times$3 & 1/4 & 1/1 & \checkmark / \checkmark & 16\\
conv\_last & 8$ \times$8 & 1/1 & 0/0 & - / - & 2\\

\hline
\multicolumn{6}{l}{$ ^{\dagger} $Zero padding}\\
\multicolumn{6}{l}{$^*$X and Y represent the horizontal and vertical axes}\\
\end{tabular}
}
\end{center}
\label{tab:vgg_720p}
\end{table}

\begin{table}[htbp]
\caption{Architecture of VGG-1080p.}
\begin{center}
\resizebox{\linewidth}{!}{
\begin{tabular}{ l c c c c c } 
\hline

Layer & Kernel & Stride & Pad$ ^{\dagger} $ (X/Y)$ ^* $ & Max Pool (X/Y) & Channels\\

\hline
\hline

conv1\_1 & 3$ \times$3 & 2/1 & 0/0 & - / - & 8\\
conv1\_2 & 3$ \times$3 & 1/2 & 0/0 & \checkmark / - & 8\\
conv2\_1 & 3$ \times$3 & 1/1 & 0/0 & - / - & 6\\
conv2\_2 & 3$ \times$3 & 1/2 & 0/0 & - / - & 6\\
conv\_last & 8$ \times$8 & 1/1 & 0/0 & - / - & 2\\

\hline
\multicolumn{6}{l}{$ ^{\dagger} $Zero padding}\\
\multicolumn{6}{l}{$^*$X and Y represent the horizontal and vertical axes}\\
\end{tabular}
}
\end{center}
\label{tab:vgg_1080p}
\end{table}

Similarly, the study in \cite{8657776} proposes an architecture with 6 convolutional layers for the same problem of generating a crowd heatmap from high-resolution images. The study in \cite{TRIANTAFYLLIDOU201865} proposes lightweight FCNs for face detection with 7 convolutional layers and 76K parameters, for facial parts detection (such as eyes, nose and mouth) with 4 convolutional layers and 20K parameters, and for combined face and parts detection with 9 convolutional layers and 101K parameters.

\textit{You only look twice (YOLT) \cite{van2018you}} is a method that detects objects of different scales in DigitalGlobe satellite images which have a size of over 250 megapixels. The architecture of YOLT is based on the YOLO architecture \cite{redmon2016you}, however, it reduces the number of layers from the original 30 down to 22. Furthermore, YOLT trains two separate models: one which processes images that correspond to areas of 200$ \times $200m$^2$ for detecting relatively small objects such as cars, airplanes, boats and buildings; and another which processes images that correspond to areas of 2500$ \times $2500m$^2$ for detecting large objects such as airports. YOLT has an inference speed of 32km$^2$/min for the former model and 6000km$^2$/min for the latter on an Nvidia Titan X GPU. 

Fast ScanNet\cite{8604098} converts VGG16 \cite{simonyan2014very} to a fully convolutional network by replacing the last fully-connected layers in VGG16 with convolutional layers of kernel size 1$ \times $1. Fast ScanNet is applied to patches of size 2800$ \times $2800 pixels, a size which is dictated by GPU memory limitations, taken from WSIs, which have $ \sim $400 patches on average. It takes about one minute for Fast ScanNet to process a WSI on a workstation with 8$ \times $Nvidia Titan X GPUs.

\begin{figure}[htbp]
\centerline{\includegraphics[width=0.48\textwidth]{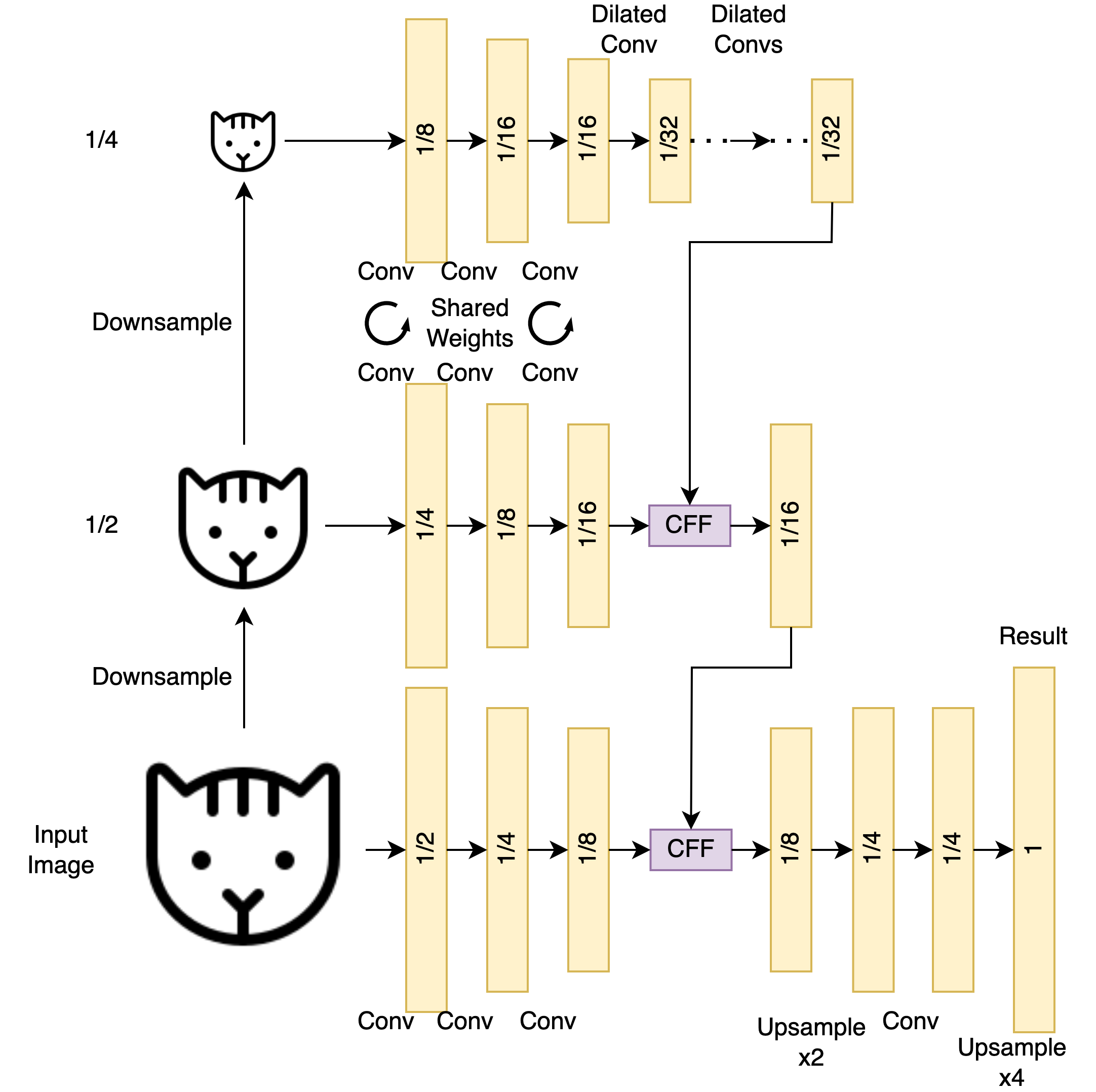}}
\caption{ICNet architecture. CFF blocks perform the fusion operation and consist of convolution and upsample layers. CFF blocks get supervision signals using downsampled annotations during the training process.}
\label{fig:ICNet}
\end{figure}

ICNet \cite{Zhao_2018_ECCV} takes advantage of both the efficiency of processing lower resolutions and the accuracy of processing higher ones by uniformly downsampling the input image to two smaller scales, processing each scale separately, and fusing the result of processing lower resolutions with higher ones. Lower resolutions are processed with more convolution layers and higher resolutions with less, which makes the entire architecture efficient, as shown in Figure \ref{fig:ICNet}. In addition, some of the layers share weights in order to increase the efficiency. ICNet is able to perform semantic segmentation on 2048$\times$1024 images at 30 frames per second with high accuracy on a Titan X GPU. Even though ICNet does not obtain state-of-the-art accuracy, it is $\sim 15 \times$ faster than methods with similar performance.

ESPNet \cite{Mehta_2018_ECCV} relies on \textit{efficient spatial pyramid (ESP)} modules which reduce the amount of computation by decomposing standard convolutions with $n \times n$ kernels into two steps. The first step applies a 1$\times$1 convolution to project feature maps with dimension $N$ to feature maps with dimension $\frac{N}{K}$. The second step applies $K$ dilated convolutions with kernel size $n \times n$ and dilation rates $2^{k-1}, k \in \{1, \dots, K\}$ to the new feature maps simultaneously, and combines the results. Concatenating the outputs of dilated convolutions creates checkerboard artifacts, therefore, a simple solution is used where the outputs of dilated convolutions are hierarchically added to each other before concatenation. ESPNet can perform semantic segmentation on 2048$\times$1024 images at 54 frames per second with an accuracy comparable to the state-of-the-art.

Neural architecture search (NAS) techniques can be used for designing better LSNs. Since LSNs need to be lightweight and contain few layers and parameters, the search space is relatively small, making NAS easier. HR-NAS \cite{Ding_2021_CVPR} is one such method that searches for network architectures that can contain both convolutions and lightweight Transformers, and may have parallel branches. HR-NAS obtains state-of-the-art results in the trade-off between efficiency and accuracy in semantic segmentation, human pose estimation and 3D object detection tasks with high-resolution inputs.

\subsection{Task-Oriented Input Compression}
\label{sec:toic}

\textit{Task-oriented input compression (TOIC)} methods compress the high-resolution inputs into lightweight representations. These representations are then given to the task DNN as input instead of the high-resolution images or videos. The exact nature of the lightweight representations and the compression procedure varies from method to method and is often highly dependent on the underlying task.

There is an important distinction between this approach and \textit{neural image compression} methods such as SlimCAE \cite{Yang_2021_CVPR}. The goal of neural image compression is to learn optimal compression algorithms for the task at hand, in order to reduce the size of stored or transmitted data. Therefore, the network that compresses and decompresses this data may be very large and inefficient. Moreover, neural image compression aims to reconstruct the input from the compressed representations, whereas TOIC does not reconstruct the input data and strives to extract compact representations that are suitable for the second part of the network which is responsible for performing the task.

Slide Graph \cite{Lu_2020_CVPR_Workshops} recognizes the loss of visual context that comes with using the cutting into patches method, and fixes this issue by building and processing a compact graph representation of the cellular architecture in breast cancer WSIs in order to predict the status of human epidermal growth factor receptor 2 (HER2) and progesterone receptor (PR), which are proteins that promote the growth of cancer cells. Slide Graph has four stages: The first stage uses a HoVer-Net \cite{graham2019hover}, which is a CNN for segmentation and classification of cellular nuclei, trained on the PanNuke dataset \cite{gamper2019pannuke} to extract features of the tissue cells. The second stage uses agglomerative clustering \cite{mullner2011modern} to group neighboring nuclei to further reduce the computational cost. The third stage constructs a graph where each vertex corresponds to a cluster and contains features extracted in the previous stage. Graph edges are constructed based on Delauney triangulation where vertices are represented by the geometric center of their corresponding cluster, which results in a planar graph. In the final stage, HER2 and PR status predictions are obtained from the constructed graph using a graph convolutional network (GCN) \cite{DBLP:conf/iclr/KipfW17}. Slide graph is more accurate than state-of-the-art methods and reduces the average inference time from 1.2 seconds of the baseline down to 0.4 milliseconds. However, these measurements do not include the graph construction phase. Therefore, the end-to-end improvement in efficiency obtained by Slide Graph is unclear.

\begin{figure}[htbp]
\centerline{\includegraphics[width=0.48\textwidth]{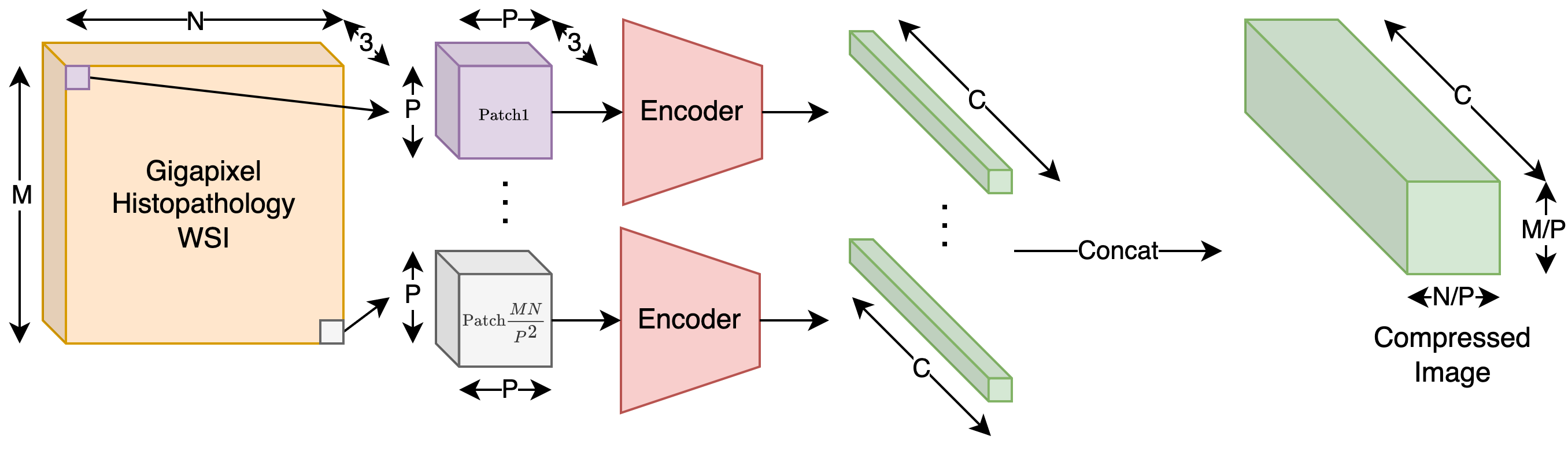}}
\caption{A method based on neural image compression for gigapixel histopathology images.}
\label{fig:NIC}
\end{figure}

The method in \cite{8809829}, shown in Figure \ref{fig:NIC}, compresses gigapixel histopathology WSIs down to a size that can be processed with a CNN on a single GPU. This compression is obtained by training an autoencoder (either VAE \cite{kingma2013auto} or bidirectional GAN \cite{donahue2016adversarial}) on image patches of size $ P \times P \times 3 $. The WSI image of size $ M \times N \times 3$ is then cut into patches of the aforementioned size, and compressed embeddings of size $ 1\times 1 \times C $ are obtained from the patches using the encoder part of the autoencoder. These embeddings are then concatenated to form a compressed image of size $ \lceil \frac{M}{P} \rceil \times \lceil \frac{N}{P} \rceil \times C $, which can be given as input to the CNN. In experiments where $ M = N = 50,000 $ and $ P = C = 128 $, the input size is reduced by a factor of $ \sim $43.

\begin{figure}[htbp]
\centerline{\includegraphics[width=0.3\textwidth]{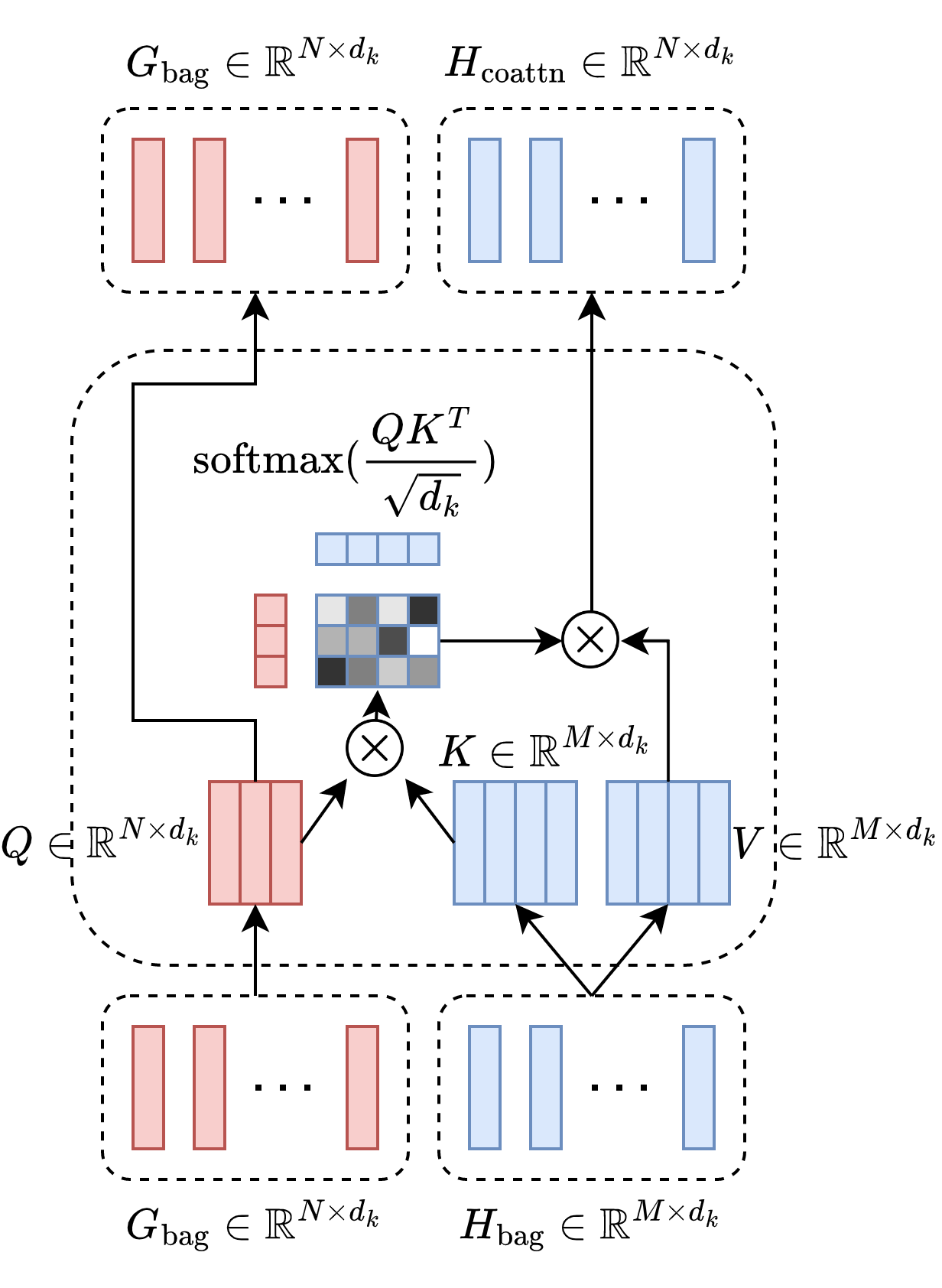}}
\caption{Genomic-Guided Co-Attention (GCA) layer.}
\label{fig:GCA}
\end{figure}

MCAT \cite{Chen_2021_ICCV} uses a combination of WSIs and genomics data for cancer survival outcome prediction. At the core of MCAT is the \textit{Genomic-Guided Co-Attention (GCA)} layer which reduces the spatial complexity of processing WSIs. MCAT processes the input in data structures known as \textit{bags}, which are unordered sets of objects of varying size without individual labels. MCAT constructs one bag ($ H_{\text{bag}} $) from multiple WSIs in order to utilize the entire tissue microenvironment, and another bag ($ G_{\text{bag}} $) from genomic features. $ H_{\text{bag}} $ is constructed by cutting the WSIs into non-overlapping $ 256 \times 256 $ pixel patches and processing each patch with a ResNet50 CNN \cite{he2016deep} pre-trained on the ImageNet dataset \cite{5206848} to obtain $ d_k $-dimensional feature embeddings. $ G_{\text{bag}} $ is constructed by categorizing genes into N different sets based on similarity and applying a fully-connected (FC) layer to obtain genomic embeddings. GCA then takes these two bags as input and performs the co-attention operation by
\begin{eqnarray}
\text{CoAttn}_{G \rightarrow H}(G, H) &=& \text{softmax} \left(\frac{QK^T}{\sqrt{d_k}} \right)V  \\ \nonumber 
&=&\text{softmax} \left(\frac{W_qGH^TW_k^T}{\sqrt{d_k}} \right)W_vH,
\label{eq:cross_attention}
\end{eqnarray}
where $ Q = W_q G $ is the query matrix, $ K = W_k H $ is the key matrix, $ V = W_v H $ is the value matrix, and $ W_q, W_k, W_v  \in \mathbb{R}^{d_k \times d_k} $ are trainable weights. The output of this operation, as shown in Figure \ref{fig:GCA}, has a dimension of $ N \times d_k $. Therefore, the subsequent self-attention layers in the MCAT network are quadratic with respect to $ N $ instead of $ M $. Since on average $ M = 15,231 $ and $ N = 6 $, this results in a massive reduction in complexity.

A subcategory of TOIC methods are \textit{frequency-domain DNNs}, which convert input RGB pixels to frequency domain representations with the help of operations such as discrete cosine transform (DCT) or wavelet transform. The intuition behind this approach is that the first few layers in CNNs often learn filters that resemble such transforms. Therefore, not only are image representations more compact in the frequency domain, but also a lower number of layers is required for processing such representations. 

\begin{figure}
\begin{center}
\begin{tabular}{ c c }
\includegraphics[width=.22\textwidth]{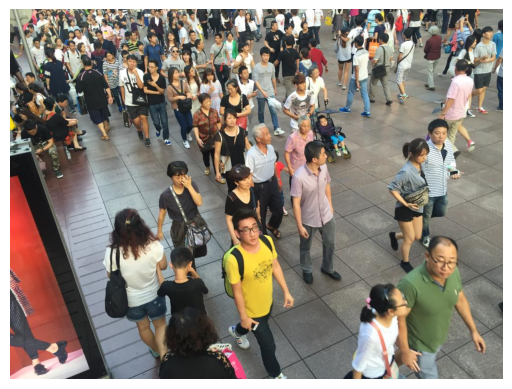}&
\includegraphics[width=.22\textwidth]{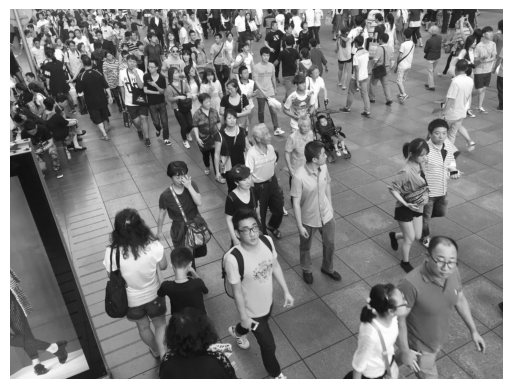}\\
(a)&(b)\\
\includegraphics[width=.22\textwidth]{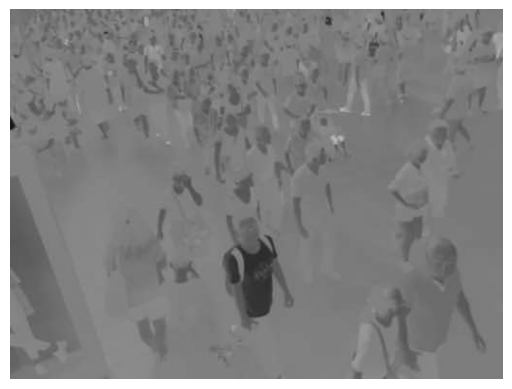}&
\includegraphics[width=.22\textwidth]{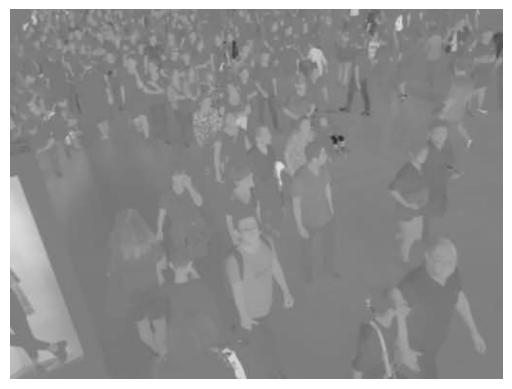}\\
(c)&(d)
\end{tabular}
\end{center}
\caption{(a) Original color image, taken from the Shanghai Tech Part B dataset \cite{zhang2016single}; (b) luma component Y, which is essentially a grayscale version of the color image; (c) chroma component Cb; and (d) chroma component Cr.}
\label{fig:ycbcr}
\end{figure}

\begin{figure}[htbp]
\centerline{\includegraphics[width=0.48\textwidth]{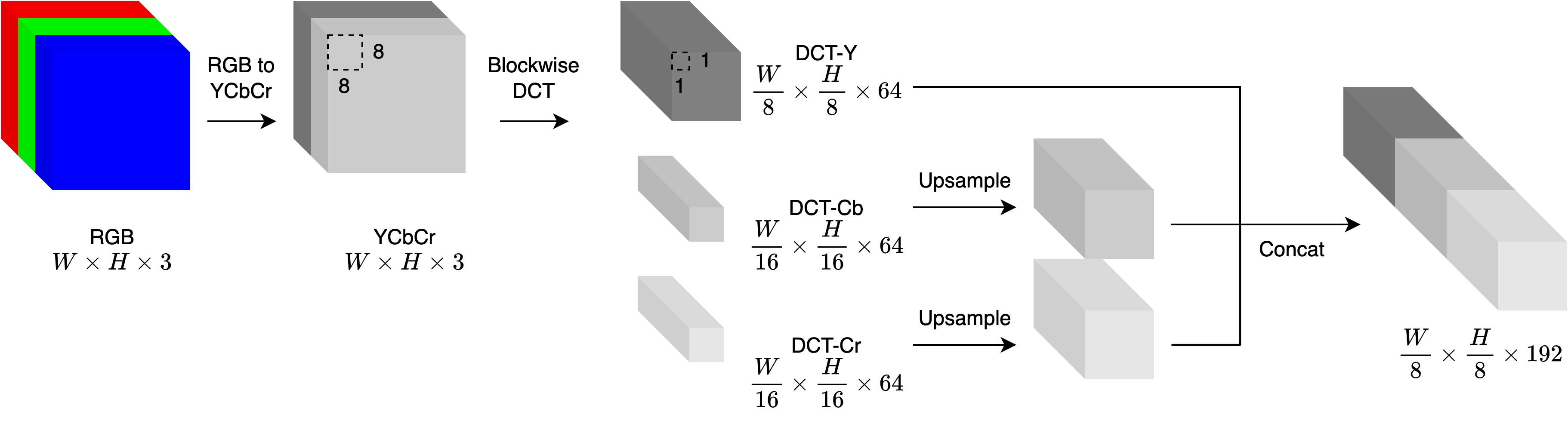}}
\caption{Initial stages of JPEG encoding, used by \cite{gueguen2018faster} to obtain frequency-domains representations of the RGB input.}
\label{fig:JPEG_DCT}
\end{figure}

The method in \cite{gueguen2018faster} uses the DCT coefficients obtained in the middle of JPEG encoding as inputs to a modified ResNet50 CNN \cite{he2016deep} for the image classification task. JPEG encoding consists of three stages. The first stage converts the input 3-channel 24-bit RGB image to the YCbCr color space by 
\begin{equation}
\begin{bmatrix}
Y\\
Cb\\
Cr
\end{bmatrix}
=
\begin{bmatrix}
0.299 & 0.587 & 0.114\\
-0.168935 & -0.331665 & 0.50059\\
0.499813 & -0.418531 & -0.081282
\end{bmatrix}
\begin{bmatrix}
R\\
G\\
B
\end{bmatrix}.
\label{eq:rgb2ycbcr}
\end{equation}
The \textit{luma} component (Y) represents the brightness, and the \textit{chroma} components (Cb and Cr) represent color. The resolution of chroma components is then reduced by a factor of 2 due to the fact that the human eye is less sensitive to fine color detail than fine brightness. Figure \ref{fig:ycbcr} shows an example image and its corresponding Y, Cb and Cr components. The second stage is a blockwise DCT, where each of the three components is partitioned into $ 8 \times 8 $ blocks that undergo a 2D DCT. The amplitude values of the frequency domain are the input representations used by this method. The DCT representations of Cb and Cr are upsampled by a factor of two and concatenated with the DCT representation of Y before being given as input to the task DNN, as shown in Figure \ref{fig:JPEG_DCT}. The rest of the JPEG encoding process contains the quantization of these representations as well as lossless compression techniques such as Huffman coding. However, this method uses the representations obtained before quantization and lossless compression.

With the help of these input representations, this method obtains DNNs that are both more accurate and up to $ 1.77 \times $ faster than ResNet50. Moreover, \cite{gueguen2018faster} includes experiments attempting to learn convolutions behaving like DCT, however, they find that this learned DCT transform leads to higher error compared to the conventional DCT transform.

The method in \cite{xu2020learning} uses the same idea for image classification and semantic segmentation tasks using ResNet50 and MobileNetV2 architectures. However, this method also prunes the 192 DCT channels with the help of a gating module that generates a binary decision for each channel. Furthermore, this study discovers that some channels are consistently pruned regardless of the particular task, and develops a static frequency channel selection scheme based on these results. This scheme prunes up to 87.5\% of the channels with little accuracy drop, if any. The method in \cite{https://doi.org/10.1049/ipr2.12466} uses the same approach for image classification, however, it uses several variants of \textit{discrete wavelet transform (DWT)} instead of DCT. The advantage of DWT over DCT is that it can obtain a better compression ratio without loss of information, however, it is more computationally expensive \cite{katharotiya2011comparative}. Experiments show that using DWT instead of DCT can lead to higher accuracy, however, the impact of DWT on inference time is unclear.

Finally, similar to images, DNNs can directly process the compressed representations obtained by video compression formats. MMNet \cite{Wang_2019_ICCV} performs efficient object detection on H.264/MPEG-4 Part 10 compressed videos \cite{richardson2004h}, one of the most commonly used video compression formats, by taking advantage of the motion information already embedded
in the video compression format. It only runs the complete feature extractor DNN on few reference frames in the video and aggregates the visual information from the subsequent frames with the help of an LSTM \cite{hochreiter1997long}. H.264 has two types of frames: \textit{I-frames} which contain a complete image, and \textit{P-frames}, also known as delta frames, which store the offset to previous frames using \textit{motion vectors} and \textit{residual errors}. In MMNet, the extracted motion vectors and residual errors for each P-frame following an I-frame are passed on to the LSTM. MMNet is $ 3 \times $ to $ 10 \times $ faster than competing models with minor loss in accuracy.

\subsection{High-Resolution Vision Transformers}

As previously mentioned, the self-attention operation in Transformers has a high complexity that increases in a quadratic fashion with respect to the number of input tokens. This operation is formulated by
\begin{equation}
Z = \mathit{softmax} \left(\frac{QK^T}{\sqrt{d_k}} \right)V,
\label{self_attention}
\end{equation}
where query $ Q = XW^Q \in \mathbb{R}^{n \times d_q} $, key $ K = XW^K \in \mathbb{R}^{n \times d_k} $ and value $ V = XW^V \in \mathbb{R}^{n \times d_v} $ are obtained from sequence of input tokens $ X = (x_1, \dots, x_n) \in \mathbb{R}^{n \times d} $, and $ W^Q $, $ W^K $ and $ W^V $ are learnable weight matrices. Due to this quadratic complexity, naive approaches, such as ViT \cite{dosovitskiy2021an}, that create a long sequence of input tokens from a high-resolution image will lead to massive complexity. On the other hand, if $ X $ contains few tokens, each input token represents a large area of the original image, leading to loss of detailed information that might be crucial to some applications.

\textit{Vision Longformer (ViL)} \cite{Zhang_2021_ICCV} is a variant of Longformer \cite{beltagy2020longformer} which has a linear complexity with respect to the number of input tokens, and is capable of processing high-resolution images. This linear complexity is achieved by adding $ n_g $ global tokens, which include the classification token $ \textit{cls} $, that serve as global memory by attending to all input tokens. Input tokens are only allowed to attend to the global tokens as well as their neighbors within a 2D window. If the number of input tokens are $ n_l $ and the 2D window size is $ w $, then the memory complexity is $ \mathcal{O}(n_g(n_g + n_l) + n_l w^2) $. When $ n_g \ll n_l $, the complexity is significantly reduced from the original $ n_l^2 $ in ViT. By using ViL in a multi-scale architecture, multi-scale Vision Longformer is able to obtain superior performance compared to the state-of-the-art in image classification, object detection and semantic segmentation while requiring less computation in terms of FLOPs in some cases.

\begin{figure}[htbp]
\centerline{\includegraphics[width=0.48\textwidth]{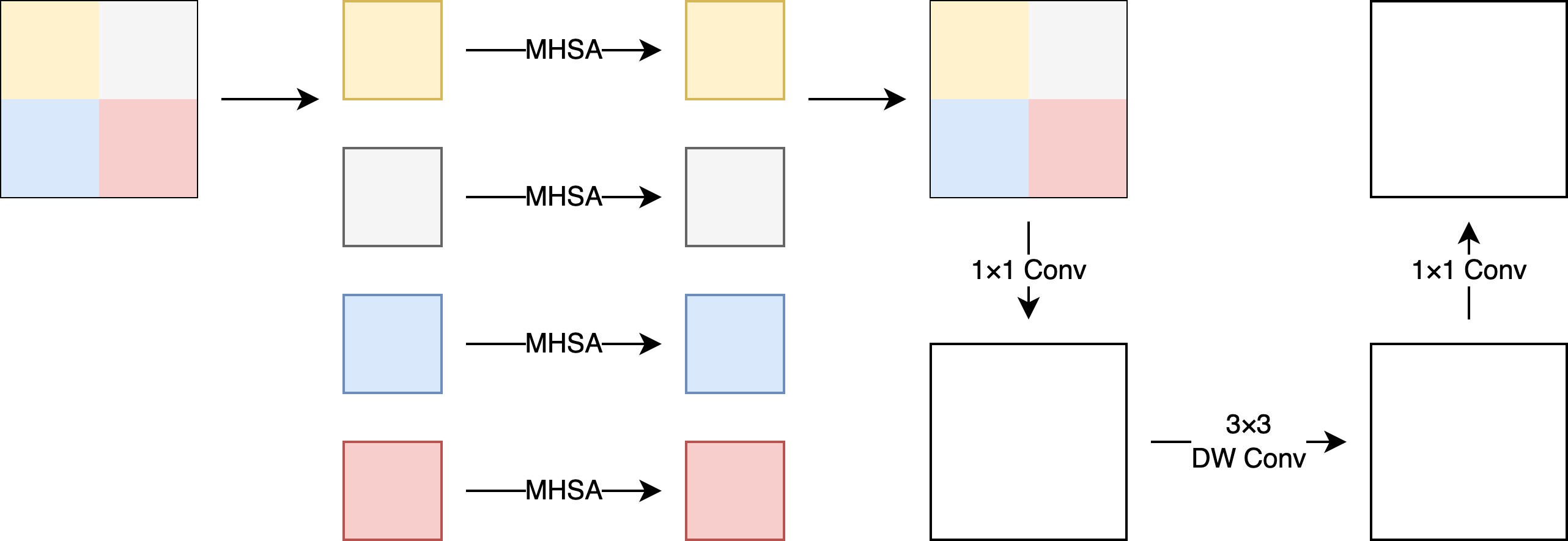}}
\caption{HRFormer block. Multi-head self-attention (MHSA) is applied only within each patch. The patches are then concatenated and followed by a depth-wise (DW) convolution.}
\label{fig:HRFormer}
\end{figure}

\textit{High-Resolution Transformer (HRFormer)} \cite{NEURIPS2021_3bbfdde8} reduces the computational complexity of self-attention by partitioning the input representations into non-overlapping patches, and performing the self-attention only within each patch. Figure \ref{fig:HRFormer} shows the building block of HRFormer, which contains a depth-wise convolution that facilitates information exchange between patches. By utilizing this augmented self-attention in a multi-scale architecture, HRFormer obtains superior performance in human pose estimation and semantic segmentation with fewer parameters and FLOPs.

\begin{figure}[htbp]
\centerline{\includegraphics[width=0.3\textwidth]{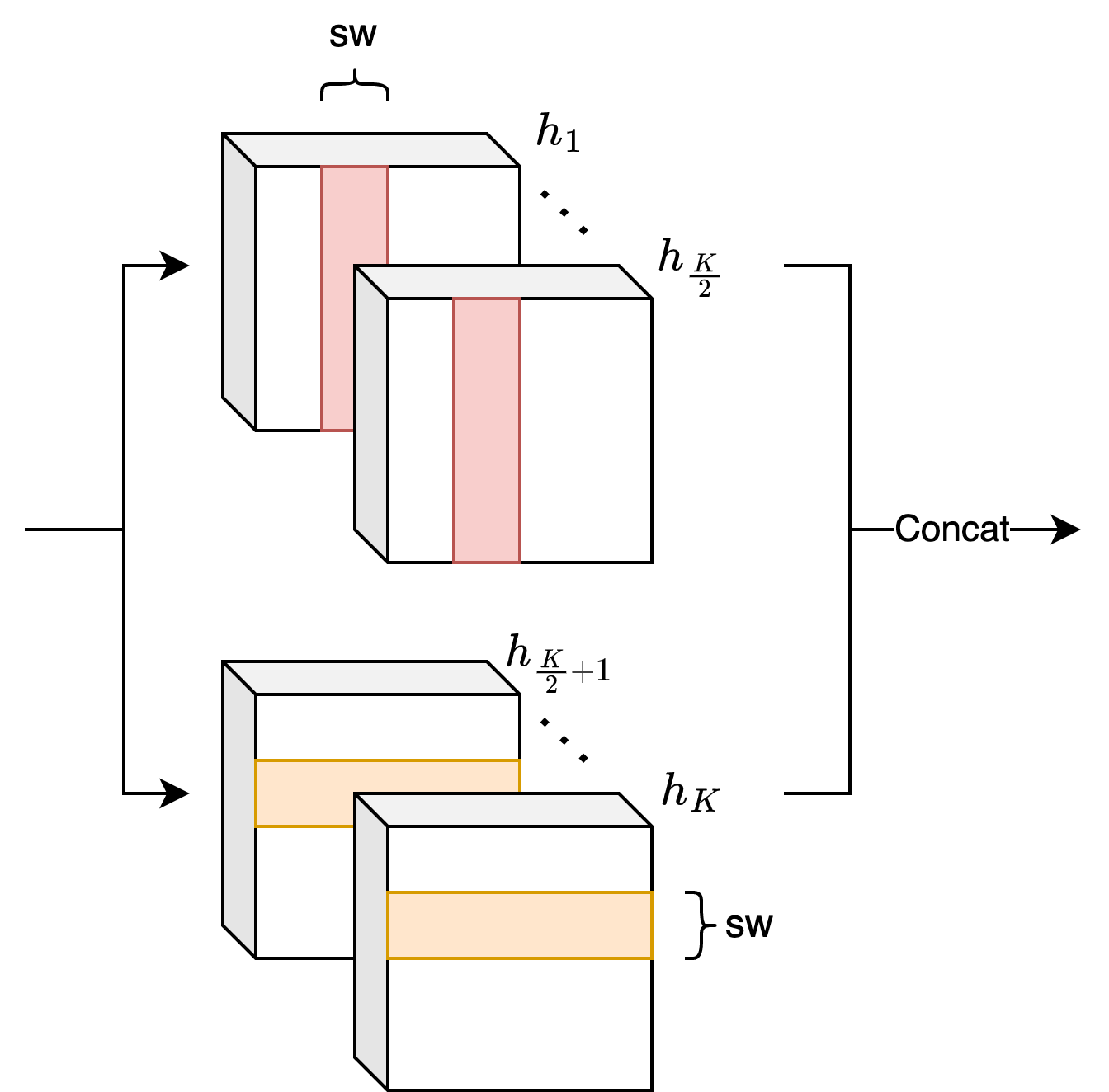}}
\caption{Cross-shaped self-attention.}
\label{fig:cross_shaped_sa}
\end{figure}

\textit{Multi-Scale High-Resolution Vision Transformer (HRViT)} \cite{Gu_2022_CVPR} uses cross-shaped self-attention \cite{Dong_2022_CVPR} and parameter sharing to decrease the computational cost of self-attention. Cross-shaped self-attention, shown in Figure \ref{fig:cross_shaped_sa}, splits the $ K $ self-attention heads present in multi-head attention into two groups: $ \{ h_1, \dots, h_{\frac{K}{2}} \} $ and $ \{ h_{\frac{K}{2}+1}, \dots, h_K \} $. These groups perform self-attention in horizontal and vertical strips in parallel. Strip width $ \text{sw} $ can be adjusted to achieve a trade-off between efficiency and performance. The linear projections for key and value tensors are shared in HRViT's blocks to save in computation and parameters. In addition to efficient self-attention, HRViT employs a convolutional stem to reduce the spatial dimension of the input. HRViT achieves the best performance-efficiency trade-off compared to state-of-the-art models for semantic segmentation.

\begin{figure}[htbp]
\centerline{\includegraphics[width=0.48\textwidth]{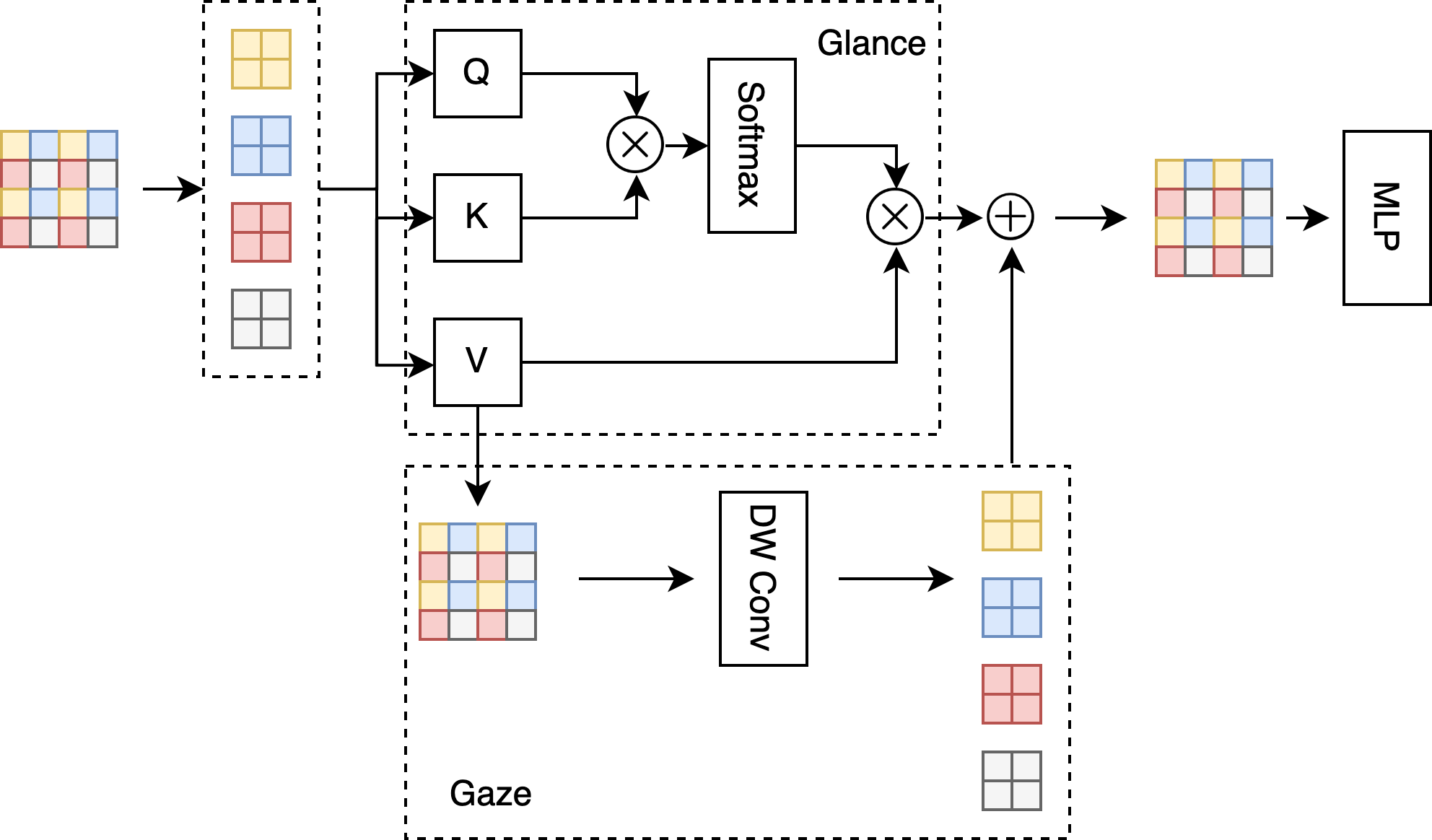}}
\caption{GG-Transformer block.}
\label{fig:ggt}
\end{figure}

Instead of restricting self-attention to patches that are neighbors in the 2D grid, \textit{Glance and Gaze Transformer (GG-Transformer)} \cite{NEURIPS2021_6c524f9d}, shown in Figure \ref{fig:ggt}, performs the self-attention within dilated partitions. Since these dilations create holes in the receptive field, a parallel branch containing depth-wise convoluion is added to compensate for the local interactions with negligible cost. GG-Transformer achieves superior performance in image classification, object detection and semantic segmentation and reduces the parameters or FLOPs in some cases.

\begin{figure}[htbp]
\centerline{\includegraphics[width=0.48\textwidth]{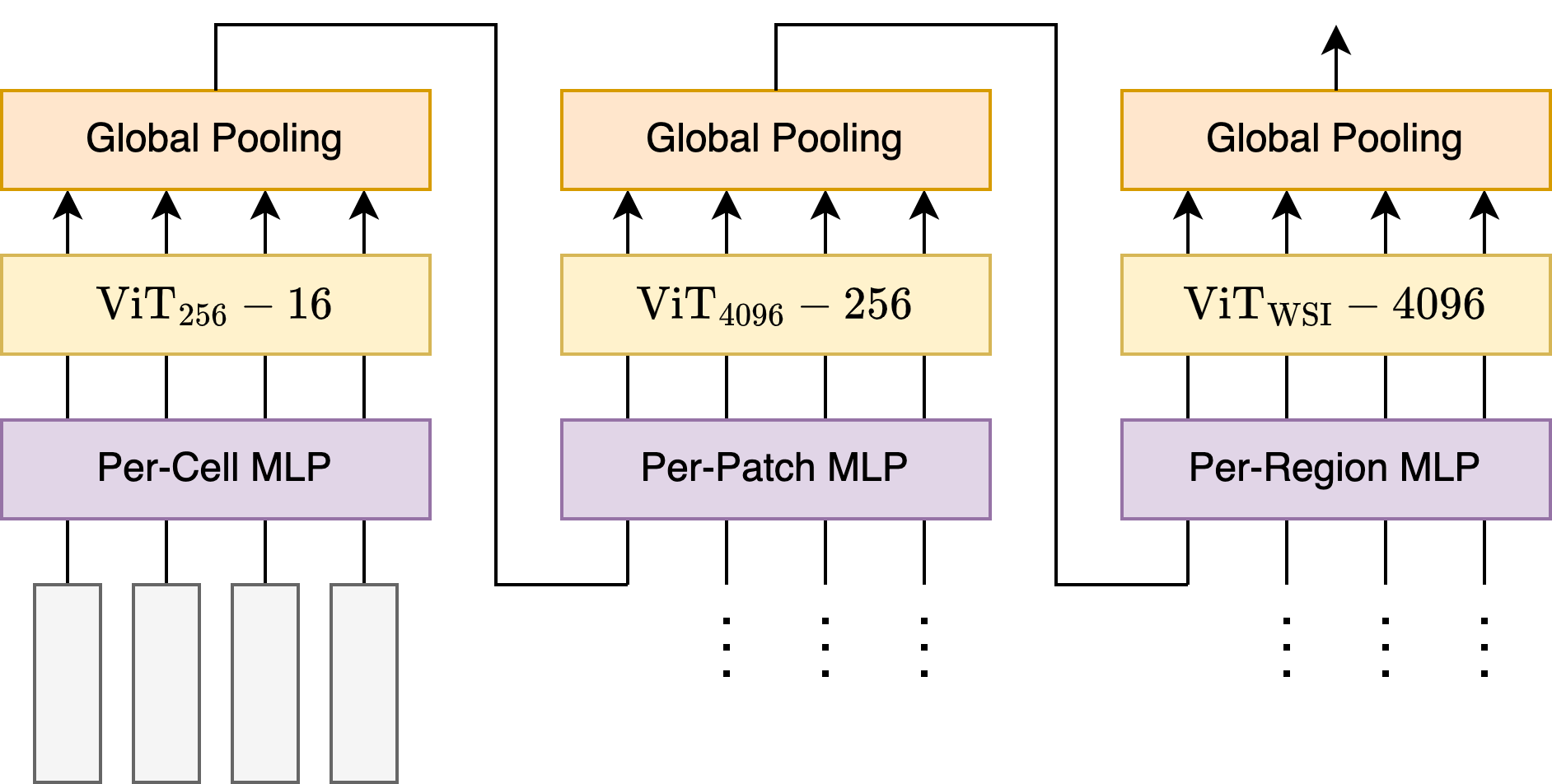}}
\caption{Hierarchical Image Pyramid Transformer (HIPT). The notation $ \text{ViT}_L - l $ means a Vision Transformer that operates on size $ L \times L$ with patch size of $ l \times l $. $ \text{ViT}_\text{WSI} $ operates on the entire WSI.}
\label{fig:hipt}
\end{figure}

\textit{Hierarchical Image Pyramid Transformer (HIPT)} \cite{Chen_2022_CVPR} processes gigapixel WSIs for the task of cancer subtyping and survival prediction. Since the input WSIs are as large as 150,000$ \times $150,000 pixels, processing them with a normal ViT and small patch size, such as 16$ \times $16, results in a massive number of parameters and computational cost requirements, and using large patch sizes  such as 4096$ \times $4096 pixels directly would result in loss of cellular information. Therefore, HIPT takes a hierarchical approach, shown in Figure \ref{fig:hipt}, where an initial ViT processes patches of 16$ \times $16 in an area of size 256$ \times $256 pixels. A second ViT then takes the aggregated tokens from the previous ViT and processes an area of size 4096$ \times $4096 pixels. A final ViT takes the aggregated tokens from the second ViT and processes the entire image.

\section{High-Resolution Datasets}\label{sec:datasets}
Table \ref{tab:high_res_datasets} lists popular datasets used in the high-resolution deep learning literature and provides information about their attributes, such as which is the deep learning application they have been primarily used for, the number of images/videos in the dataset and their resolution, the type of available annotations, whether they specify training/validation/test set splits, the year of publication, and whether they are publicly available. 
It is important to note that studies reported in some papers create customized datasets. For instance, \cite{Gao_2018_CVPR} constructs a dataset from YFCC100M \cite{thomee2016yfcc100m}; \cite{Tzelepi2020} constructs datasets from AFLW \cite{koestinger11a}, MTFL \cite{zhang2014facial} and WIDER FACE \cite{yang2016wider}; and \cite{van2018you} constructs datasets from DigitalGlobe satellites, Planet satellites, and aerial platforms. 

The Cancer Genome Atlas (TCGA) program is a collaboration between National Cancer Institute (NCI) and National Human Genome Research (NHGRI)\footnote{\url{https://www.cancer.gov/about-nci/organization/ccg/research/structural-genomics/tcga}}. Since 2006, TCGA has generated over 2.5 petabytes of publicly available data which has led to improvements in cancer diagnosis, treatment, and prevention. Among efficient high-resolution deep learning methods, the most widely used subset of this data is the breast invasive carcinoma (BRCA), which is outlined in Table \ref{tab:high_res_datasets}. However, TCGA provides data for many other types of cancer, such as bladder urothelial carcinoma (BLCA), glioblastoma and lower grade glioma (GBMLGG), lung adenocarcinoma (LUAD), and uterine corpus endometrial carcinoma (UCEC). These are used in some studies, and have properties similar to that of BRCA.

\begin{table*}[htbp]
\caption{List of Popular High-Resolution Datasets}
\begin{center}
\resizebox{\linewidth}{!}{
\begin{tabular}{ l l l l l l c l } 
\hline
Name & Applications & Resolution (Pixels) & \# of Samples & Annotations & Splits & Year & Availability\\
\hline
\hline

Supervisely Persons$^\ddagger$ & Person Segmentation & 800$ \times $1116 to 9933$ \times $6622 & 5711 images & Pixel Mask & None & 2018 & Public\\

PANDA\cite{wang2020panda} & Person Detection & $ > $25K$ \times $14K & 555 frames$^{\mathsection}$ & Person Bounding Box & None & 2020 & Upon Request\\

UCF\_CC\_50\cite{idrees2013multi} & Crowd Counting & 2888$ \times $2101 on average & 50 images & Head Annotations$^{*}$ & None & 2013 & Public\\

Shanghai Tech Part A\cite{7780439} & Crowd Counting & 868$ \times $589 & 482 images & Head Annotations & Train \& Test & 2016 & Public\\

Shanghai Tech Part B\cite{7780439} & Crowd Counting & 1024$ \times $768 & 716 images & Head Annotations & Train \& Test & 2016 & Public\\

UCF-QNRF\cite{idrees2018composition} & Crowd Counting & 2902$ \times $2013 on average & 1535 images & Head Annotations & Train \& Test & 2018 & Public\\

PANDA Crowd\cite{wang2020panda} & Crowd Counting & 25,151$ \times $14,151 to 26,908$ \times $15,024 & 45 images & Person Bounding Box & None & 2020 & Upon Request\\

JHU-CROWD++\cite{sindagi2020jhu-crowd} & Crowd Counting & 1430$ \times $910 on average & 4372 images & Head Annotations & Train, Val \& Test & 2020 & Public\\

NWPU-Crowd\cite{wang2020nwpu} & Crowd Counting & 3209$ \times $2191 on average & 5109 images & Head Annotations & Train, Val \& Test & 2020 & Public\\

DISCO\cite{hu2020ambient} & Audio-Visual Crowd Counting & 1920$ \times $1080 (Full HD) & 1935 images & Head Annotations & Train \& Test & 2020 & Public\\

CityScapes\cite{Cordts_2016_CVPR} & Autonomous Driving & 2048$ \times $1024 & 5K images & Pixel Mask & Train, Val \& Test & 2016 & Upon Request\\

SYNTHIA-RAND\cite{Ros_2016_CVPR} & Autonomous Driving & 1280$ \times $720 (HD) & $\sim$13K images & Pixel Mask & Train \& Test & 2016 & Public\\

ApolloScape\cite{8753527} & Autonomous Driving & 3384$ \times $2710 & $\sim$113K images & Pixel Mask & Train \& Test & 2020 & Upon Request\\

Argoverse-HD\cite{10.1007/978-3-030-58536-5_28} & Autonomous Driving & 1920$ \times $1200 & 89 videos & Bounding Box & Train, Val \& Test & 2020 & Public\\

BDD100K\cite{Yu_2020_CVPR} & Autonomous Driving & 1280$ \times $720 (HD) & 100K videos & Bounding Box & Train, Val \& Test & 2020 & Upon Request\\

PASCAL-Context\cite{mottaghi_cvpr14} & Scene Understanding & 500$ \times $375 to 500$ \times $500 & 10,103 images & Pixel Mask & Train \& Test & 2014 & Public\\

ADE20K\cite{zhou2017scene} & Scene Understanding & 683$ \times $512 to 2100$ \times $2100 & 27,574 images & Pixel Mask & Train \& Test & 2017 & Upon Request\\

COCO-Stuff 10K\cite{caesar2018coco} & Scene Understanding & $ \sim $640$ \times $480 & 10K images & Pixel Mask & Train \& Test & 2018 & Public\\

DeepGlobe\cite{demir2018deepglobe} & Land Cover Classification & 2448$ \times $2448 & 1146 images & Pixel Mask & Train, Val \& Test & 2018 & Public\\

Copernicus\cite{buchhorn2020copernicus} & Land Cover Classification & 20,160$\times$20,160 & 94 images & Pixel Mask & None & 2015-2019 & Public\\

fMoW\cite{fmow2018} & Aerial Image Classification & up to 16,032$ \times $14,840 & 1,047,691 images & Classes & Train, Val \& Test & 2018 & Public\\

KID\cite{Koulaouzidis2017} & Capsule Endoscopy & 360$ \times $360 & $\sim$2500 frames & Pixel Mask & None & 2017 & Public (N/A)\\

CAD-CAP\cite{Leenhardt2020-hv} & Capsule Endoscopy & 576$ \times $576 & 25,124 frames & Pixel Mask & Train \& Test & 2020 & Upon Request\\

CAMELYON16\cite{yang2016wider} & Pathology & up to 200,000$ \times $100,000 & 400 images & Pixel Mask & Train \& Test & 2016 & Public\\

TUPAC16\cite{VETA2019111} & Pathology & $ \sim $50,000$ \times $50,000 & 821 images & Proliferation Score$^{\dagger}$ & Train \& Test & 2016 & Public\\

BACH Part B\cite{ARESTA2019122} & Pathology & (39,980-62,952)$ \times $(27,972-44,889) & 40 images & Pixel Mask & Train \& Test & 2019 & Public\\

TCGA-BRCA\cite{Koboldt2012} & Pathology & up to 150,000$ \times $100,000 & 709 images & Classes & None & 2020 & Public\\

PCa-Histo\cite{jin2021learning} & Pathology & (1968±216)$ \times $(9392±4794) & 266 images & Pixel Mask & Train, Val \& Test & 2021 & Private\\

INbreast\cite{Moreira2012} & Breast Cancer Detection & 2560$ \times $3328 to 3328$ \times $4084 & 410 images & Pixel Mask & Train \& Test & 2012 & Public\\

UA-DETRAC\cite{CVIU_UA-DETRAC} & Video Object Detection & 960$ \times $540 & 140K frames & Bounding Box & Train \& Test & 2015 & Public\\

ImageNet-VID\cite{russakovsky2015imagenet} & Video Object Detection & 176$ \times $132 to 1280$ \times $720 (HD) & 5354 videos & Bounding Box & Train, Val \& Test & 2015 & Public\\

FAIR1M\cite{SUN2022116} & Fine-Grained Object Detection & 600$\times$600 to 10,000$\times$10,000 & 40,000 images & Bounding Box & Train \& Test & 2021 & Public (N/A)\\

COCO\cite{lin2014microsoft} & \begin{tabular}{@{}l@{}}Object Detection\\Human Pose Estimation\end{tabular} & $ \sim $640$ \times $480 & $ > $200K images & \begin{tabular}{@{}l@{}}Pixel Mask\\Keypoints\end{tabular} & Train, Val \& Test & 2014 & Public\\

\hline
\multicolumn{8}{l}{$^{\mathsection}$A frame is a single image in a sequence representing a video}\\
\multicolumn{8}{l}{$^{*}$The locaion for the center of each human head in the image is specified}\\
\multicolumn{8}{l}{$^{\dagger}$A measure of the number of cells in a tumor that are dividing}\\
\multicolumn{8}{l}{$^{\ddagger}$\url{https://github.com/supervisely-ecosystem/persons}}\\
\end{tabular}
}
\end{center}
\label{tab:high_res_datasets}
\end{table*}

\section{Discussion and Open Issues}\label{sec:discussion}
Each of the approaches introduced in Section \ref{sec:methods} has its advantages and disadvantages and is useful in certain situations. NUD (Section \ref{sec:non_uniform_downsampling}) works well in cases where the salient area is small compared to the entire image, and thus, it is possible to sample many pixels from such areas. This requirement is satisfied in gaze estimation or object detection problems. Our conjecture is that it would also work well in problems such as hand gesture detection and non-cropped facial expression recognition, although these tasks are not yet explored in the literature in combination with NUD. However, when the salient area is large, for instance densely populated scenes in visual crowd counting or a scene fully covered with objects in object detection, the quality gain obtained by sampling from salient areas will be negligible, and the result of NUD will be similar that of uniform downsampling \cite{bejnordi2022salisa}.

Similarly, SZS methods (Section \ref{sec:szs}) require the salient area to be small, otherwise they zoom everywhere and save little time and computation. This also means that the effectiveness of NUD and SZS methods may vary based on the specific input. For instance, the more people there are in an image processed for crowd counting, or the more tumors there are in cancer detection, the less efficient such methods will be, unless there are specific safeguards that prevent them from performing an enormous number of computations, such as GigaDet \cite{CHEN202214} which processes at most $ K $ patch candidates.

Furthermore, NUD methods are not effective when the resulting resolution is extremely smaller compared to the input resolution, for instance, when gigapixel inputs need to be resized down to HD, as this would result in highly distorted images which makes it difficult for the task DNN to perform well. Even when the gap between the two resolutions is not extremely large, NUD can lead to high distortions in some cases, for instance, it may completely distort and change the shape of the edges of a gastrointestinal lesion, making it difficult for the task network to detect useful features. This may reduce accuracy despite the fact that more pixels are sampled from salient  areas. As explained in Section \ref{sec:non_uniform_downsampling}, some methods try to mitigate the distortion by using structured grids. However, this may limit the benefits obtained by NUD. 

In addition, since NUD is enlarging some parts of the image compared to uniform downsampling, some areas of the resulting image will be smaller than they would be with uniform downsampling. Thus, if the saliency map is not of high quality, unimportant areas will be enlarged and the ones important for the final task will shrink, resulting in accuracy loss. This is directly at odds with the requirement that the saliency detection method should be low-overhead, creating another trade-off that needs to be carefully balanced. Moreover, as explained in Section \ref{sec:non_uniform_downsampling}, some variations of NUD require an external supervision signal or regularization term to train the saliency detection network, which can be difficult to design. In NUD or SZS methods that detect saliency in videos based on the results obtained from previous frames, such as SALISA \cite{bejnordi2022salisa} and REMIX \cite{10.1145/3447993.3483274}, when the difference between subsequent frames is high, the method needs to be reset to processing the entire high-resolution image. When this occurs frequently, the obtained benefits are diminished.

As mentioned in Section \ref{sec:lsn}, LSNs need to designed, trained and well optimized for the specific problem at hand, which is not an easy task. Furthermore, since LSNs produce an output for each scanned area of the input, they are suitable for tasks where the output has the form of a map, such as dense classification or dense regression problems. Moreover, the scanning nature of LSNs means that all areas of the image are treated similarly, therefore, they are better suited for situations where there is no perspective and objects of the same type have the same size regardless of their location, such as WSIs and remote sensing, as opposed to surveillance and crowd counting where people close to the camera are larger than people far away.

Since TOIC methods extract representations that are both compressed and suitable for the task at hand, they often need to be tailored to the specific problem, which requires high domain knowledge. Both Slide Graph \cite{Lu_2020_CVPR_Workshops} and MCAT \cite{Chen_2021_ICCV} presented in Section \ref{sec:toic} are based on domain knowledge about cellular structure of tissues and biological function of genes, respectively. Almost all frequency-domain DNNs try to preserve the architecture of the CNNs they are based on. However, since the interpretation of features in frequency-domain is different, and they have certain properties such as being non-negative, it might be better to customize the architectural elements for the frequency domain, as CS-Fnet \cite{9591201} does.


Most high-resolution Vision Transformer methods try to reduce the quadratic cost of self-attention to linear, and then compensate the accuracy loss by learning data transformations using convolutions. To keep the overhead of convolutions low, depth-wise convolution is typically used. Additionally, most high-resolution ViTs utilize a multi-scale architecture in order to capture features of various scales. High-resolution ViTs are more general purpose than other high-resolution deep learning methods and are often used for a large variety of tasks.

\section{Conclusion and Outlook}\label{sec:conclusion}
Processing high-resolution images and videos with deep learning is crucial in various domains of science and technology. However, few methods exist that address the computational challenges. Among existing methods, the trend of designing solutions specifically for the problem at hand is clearly visible. This can be an issue in tasks for which high-resolution datasets are not available. Similar to model compression approaches, both modifying existing methods and designing an efficient high-resolution method from scratch are viable approaches.

Efficient high-resolution deep learning is in its infancy and there is a lot of room for improvement. For instance, a number of attention-free MLP-based methods have been recently proposed as lightweight alternatives for Transformers \cite{guo2021can}, which try to mimic the global receptive field of Transformers without the self-attention mechanism. Exploiting such architectures for efficient processing of high-resolution inputs would be an interesting research direction. Furthermore, the multimodal co-attention in MCAT \cite{Chen_2021_ICCV} can be applied to many other multimodal tasks, especially the ones with audio, vision and language modalities. Moreover, frequency-domain representations can be explored as inputs to ViTs, which can lead to more efficiency compared to frequency-domain CNNs. For instance, ViTs can take separate patches from DCT-Cb, DCT-Cr and DCT-Y components, bypassing the need to upsample DCT-Cb and DCT-Cr to match the dimensions of DCT-Y.

The combination of efficient high-resolution deep learning with other efficient deep learning methods, such as model compression \cite{8253600}, dynamic inference \cite{2102.04906}, collaborative inference \cite{Carreira_2018_ECCV} and continual inference \cite{hedegaard2022coInf}, is an unexplored area of research. For instance, if the saliency detection network is a lightweight version of the task network, NUD can be combined with early exiting, where the output of the saliency detection network would be a fast, but less accurate, early result. This is simple to implement in dense regression problems such as depth estimation and crowd counting, where the output of the task can be interpreted as a form of saliency.

Moreover, with the adoption of edge and cloud computing, transmission of high-resolution inputs to servers for processing is a real challenge. As a solution, efficient high-resolution deep learning methods can be combined with edge computing paradigms. For instance, the downsampled images in NUD and compressed representation in TOIC can be transmitted instead of the original inputs. This would be a form of split computing (also known as collaborative intelligence) \cite{matsubara2021split, bakhtiarnia2022dynamic}, where the initial portion of computation is performed on a resource-constrained end-device, and the compact intermediate representation is then transmitted to a server where the rest of the computation is carried out. A study using this idea for high-resolution images captured by drones is reported in \cite{10.1117/12.2557988}.

\bibliographystyle{IEEEtran}
\bibliography{references.bib}

\appendices

\section{Data Sources}
\label{sec:data_source}

Data Sources and details for device camera resolutions are shown in Table \ref{tab:data_sources}.

\begin{table*}
\caption{Details for device camera resolutions. All links were accessed at 26 July 2022.}
\begin{center}
\begin{tabular}{ c c c l }
    Device Camera & Year & Resolution (MP) & Source\\
    \hline
    Apple iPhone Rear Camera & 2007 & 2 & \url{https://en.wikipedia.org/wiki/IPhone_(1st_generation)}\\
    & 2008 & 2 & \url{https://en.wikipedia.org/wiki/IPhone_3G}\\
    & 2009 & 3 & \url{https://en.wikipedia.org/wiki/IPhone_3GS}\\
    & 2010 & 5 & \url{https://en.wikipedia.org/wiki/IPhone_4}\\
    & 2011 & 8 & \url{https://en.wikipedia.org/wiki/IPhone_4S}\\
    & 2012 & 8 & \url{https://en.wikipedia.org/wiki/IPhone_5}\\
    & 2013 & 8 & \url{https://en.wikipedia.org/wiki/IPhone_5S}\\
    & 2014 & 8 & \url{https://en.wikipedia.org/wiki/IPhone_6}\\
    & 2015 & 12 & \url{https://en.wikipedia.org/wiki/IPhone_6S}\\
    & 2016 & 12.2 & \url{https://en.wikipedia.org/wiki/IPhone_SE_(1st_generation)}\\
    & 2017 & 12 & \url{https://en.wikipedia.org/wiki/IPhone_X}\\
    & 2018 & 12 & \url{https://en.wikipedia.org/wiki/IPhone_XS}\\
    & 2019 & 12 & \url{https://en.wikipedia.org/wiki/IPhone_11_Pro}\\
    & 2020 & 12 & \url{https://en.wikipedia.org/wiki/IPhone_12_Pro}\\
    & 2021 & 12 & \url{https://en.wikipedia.org/wiki/IPhone_13_Pro}\\
    & 2022 & 12 & \url{https://en.wikipedia.org/wiki/IPhone_SE_(3rd_generation)}\\
    \hline
    Samsung Galaxy S Rear Camera & 2010 & 5 & \url{https://en.wikipedia.org/wiki/Samsung_Galaxy_S}\\
    & 2011 & 8 & \url{https://en.wikipedia.org/wiki/Samsung_Galaxy_S_II}\\
    & 2012 & 8 & \url{https://en.wikipedia.org/wiki/Samsung_Galaxy_S_III}\\
    & 2013 & 13 & \url{https://en.wikipedia.org/wiki/Samsung_Galaxy_S4}\\
    & 2014 & 16 & \url{https://en.wikipedia.org/wiki/Samsung_Galaxy_S5}\\
    & 2015 & 16 & \url{https://en.wikipedia.org/wiki/Samsung_Galaxy_S6}\\
    & 2016 & 12 & \url{https://en.wikipedia.org/wiki/Samsung_Galaxy_S7}\\
    & 2017 & 12 & \url{https://en.wikipedia.org/wiki/Samsung_Galaxy_S8}\\
    & 2018 & 12 & \url{https://en.wikipedia.org/wiki/Samsung_Galaxy_S9}\\
    & 2019 & 16 & \url{https://en.wikipedia.org/wiki/Samsung_Galaxy_S10}\\
    & 2020 & 108 & \url{https://en.wikipedia.org/wiki/Samsung_Galaxy_S20}\\
    & 2021 & 108 & \url{https://en.wikipedia.org/wiki/Samsung_Galaxy_S21}\\
    & 2022 & 108 & \url{https://en.wikipedia.org/wiki/Samsung_Galaxy_S22}\\
    \hline
    Microsoft HoloLens Camera & 2016 & 2.4 & \url{https://docs.microsoft.com/en-us/hololens/hololens1-hardware}\\
    & 2019 & 8 & \url{https://www.microsoft.com/en-us/hololens/hardware}\\
    \hline
    Raspberry Pi Camera & 2013 & 2.1 & \url{https://en.wikipedia.org/wiki/Raspberry_Pi#Accessories}\\
    & 2016 & 8 & \url{https://en.wikipedia.org/wiki/Raspberry_Pi#Accessories}\\
    & 2020 & 12.3 & \url{https://en.wikipedia.org/wiki/Raspberry_Pi#Accessories}\\
    \hline
    DJI Phantom Camera & 2012 & 12 & \url{https://en.wikipedia.org/wiki/GoPro#HERO3_(White/Silver/Black)}\\
    & 2013 & 14 & \url{https://www.dji.com/dk/phantom-2-vision}\\
    & 2014 & 14 & \url{https://www.dji.com/dk/phantom-2-vision-plus}\\
    & 2015 & 12.4 & \url{https://www.dji.com/dk/phantom-3-pro}\\
    & 2016 & 20 & \url{https://en.wikipedia.org/wiki/Phantom_(UAV)#Current_Phantom_drones}\\
    & 2017 & 20 & \url{https://en.wikipedia.org/wiki/Phantom_(UAV)#Current_Phantom_drones}\\
    & 2018 & 20 & \url{https://en.wikipedia.org/wiki/Phantom_(UAV)#Current_Phantom_drones}\\
\end{tabular}
\end{center}
\label{tab:data_sources}
\end{table*}

\end{document}